\documentclass[a4paper,fleqn]{cas-dc}

\usepackage[authoryear,longnamesfirst]{natbib}

\usepackage{amsmath,amssymb,amsfonts}
\usepackage{array}
\usepackage{url}
\usepackage{multirow}
\usepackage{booktabs}
\usepackage{gensymb} 

\def\mrII{\multirow{2}{*}} 
\def\mrIII{\multirow{3}{*}} 
 
\def\FL2{FAST-LIO2}
\def\FL{FAST-LIO}
\def\g{\color{gray}}
\def\bl{\textcolor{blue}}

\def\tsc#1{\csdef{#1}{\textsc{\lowercase{#1}}\xspace}}
\tsc{WGM}
\tsc{QE}

\begin{document}
\let\WriteBookmarks\relax
\def\floatpagepagefraction{1}
\def\textpagefraction{.001}

\shorttitle{Parameter Sensitivity Analysis for Aerial LiDAR-Inertial Odometries}    

\shortauthors{R. Milijas et~al.}  

\title [mode = title]{Parameter Sensitivity Analysis for Aerial LiDAR-Inertial Odometries in low-altitude flights}  


\tnotetext[1]{The work presented in this paper was funded by the European Union under the MARBLE project (GA No: 101136349) and through the National Recovery and Resiliance Plan under the grant NPOO.C3.2.R3-I1.01.0274 (project OTIP - Point cloud for industrial plant digitalization) and by MICIU/AEI/10.13039/501100011033 and ERDF, EU, through the Project "RAISE: Robots A\'ereos Inteligentes en Cooperaci\'on Estrecha con Sistemas IoT para la Inspecci\'on Avanzada de Viaductos", under Grant PID2023-149683OB-I00. Views and opinions expressed are however those of the author(s) only and do not necessarily reflect those of the European Union or the European Research Executive Agency. Neither the European Union nor the granting authority can be held responsible for them.} 

\tnotetext[2]{The authors have no conflicts of interest to disclose.} 

%

\author[1]{Robert Milijas}[orcid=0000-0002-8796-312X]

\cormark[1]

\fnmark[1]

\ead{robert.milijas at marble.eu}

\ead[url]{https://www.marble.eu}

\credit{Conceptualization, Data curation, Methodology, Validation, Writing - original draft, Writing - review \& editing}

\affiliation[1]{organization={CoE MARBLE - Centre of Excellence in Maritime Robotics and Technologies for Sustainable Blue Economy},
            addressline={Unska 3}, 
            city={Zagreb},
            postcode={10000}, 
            state={},
            country={Croatia}}

\author[2]{Jose Ramiro {Martinez-de Dios}}[orcid=0000-0001-9431-7831]

\fnmark[2]

\ead{jdedios at us.es}

\ead[url]{https://www.us.es}

\credit{Conceptualization, Writing - original draft, Writing - review \& editing, Supervision, Funding acquisition}

\affiliation[2]{organization={GRVC Robotics Laboratory, Universidad de Sevilla},
            addressline={Camino de los Descubrimientos}, 
            city={Sevilla},
            postcode={41092}, 
            state={},
            country={Spain}}

\author[1,3]{Stjepan Bogdan}[orcid=0000-0003-2636-3216]

\fnmark[3]

\ead{stjepan.bogdan at fer.unizg.hr}

\ead[url]{https://www.fer.unizg.hr}

\credit{Conceptualization, Supervision, Funding acquisition, Writing - review \& editing}

\affiliation[3]{organization={University of Zagreb Faculty of Electrical Engineering and Computing, Laboratory for Robotics and Intelligent Control Systems (LARICS)},
            addressline={Unska 3}, 
            city={Zagreb},
            postcode={10000}, 
            state={},
            country={Croatia}}
\cortext[1]{Corresponding author}

\fntext[1]{}


\begin{abstract}
LiDAR-based SLAM (Simultaneous Localization and Mapping) and LIO (LiDAR-inertial odometry) algorithms are often used for precise navigation of unmanned aerial vehicles, especially during interactions with the aerial robot's environment. However, the performance of these algorithms is greatly dependent on the scenario, LiDAR, and robot motion characteristics, often requiring an intensive tuning process to achieve the desired performance. To aid these tuning efforts, this paper analyzes the influence on performance of the parameters of an EKF-based LIO algorithm (\FL2) and the LIO module of a graph-based SLAM algorithm (Cartographer) on aerial LiDAR SLAM datasets recorded using different LiDARs in low-to-moderate-altitude flights in diverse environments. The analysis is conducted on the absolute trajectory error (ATE) resulting from processing the datasets with the LIO algorithms configured with each combination of parameters obtained in an exhaustive grid search. The relationship between individual parameters and the ATE results is assessed using Pearson's correlation, while the influence of each parameter is assessed using random forest permutation importance analyses with random forest models trained to predict the resulting ATE values based on the choice of parameters. The performed analysis obtains for Cartographer and \FL2: i) the identification of parameters with stronger influence in performance, ii) a simplified tuning procedure, and iii) tuning recommendations. Using the proposed tuning recommendations, both algorithms obtain on the analyzed datasets ATE values within 5 cm to the optimal performance found in the grid search procedure in 94\% of the analyzed cases.
\end{abstract}

\begin{keywords}
Unmanned Aerial Vehicle \sep Sensitivity analysis \sep LiDAR-Inertial Odometry \sep SLAM \sep Navigation \sep Random forests 
\end{keywords}

\maketitle

\setcitestyle{numbers,square,comma}

\section{Introduction}
\label{sec:introduction}

The use of autonomous Unmanned Aerial Vehicles (UAVs) in an increasingly broader range of applications has generated strong interest in LiDAR-based Simultaneous Localization and Mapping (SLAM) and LiDAR-inertial odometry
(LIO) algorithms. A wide variety of schemes have been developed making use of different representations, types of scan matching, robot motion estimation approaches, or ways of integrating IMU measurements, among many others \cite{Lee2024, Cadena2016}. Many existing methods have demonstrated excellent results in geometrically rich scenarios. However, LiDAR-based SLAM and LIO algorithms are known to be sensitive to the setting of the environment \cite{Koide2021_adaptive, Koide2021_autotuning}.
Characteristics such as the environment size, the distance between the LiDAR and the objects in the environment, the number and density of geometrical features, or the type of robot motion and velocity have strong impact on the performance of these algorithms.

LiDAR-based odometry and SLAM methods are known to demand extensive parameter tuning to adapt to the particularities of the environment \cite{Wu2024_ICRA, Cheng2025_RAL}. The tuning of parameters of these algorithms has become a critical step when deploying autonomous UAVs in a given scenario, often remaining prone to instability when deployed in diverse environments or when changing the sensory payload \cite{Cadena2016, Cheng2025_RAL}. In fact, in many cases parameters tuned for some environments fail in other types of environment, requiring retuning \cite{Koide2021_adaptive, Koide2021_autotuning}. Environments with strong symmetries or lack of geometrical information also pose well-known difficulties \cite{Lee2023_symmetry, Ebadi2021_DARE_degraded, julio2019}. Tuning LiDAR-based odometry and SLAM algorithms is also known to be a very time-consuming process. This is highlighted in a recent survey paper by the teams participating in the 2021 DARPA SubT challenge \cite{Ebadi2023_DARPA_SubT} where most of the teams report allocating a vast amount of human and computational resources to tuning their SLAM or LIO solutions. For example, the top performing team CERBERUS states that their SLAM solution "\textit{requires a careful finetuning of all available parameters. For example, the degeneracy detection relies on a hand-tuned set of parameters and is robot-dependent}".

Several approaches for reducing the tuning effort have been reported. Some authors employ auto-tuning and hyperparameter optimization techniques to configure the algorithms 
\cite{Koide2021_autotuning, Menezes2020}. Other approaches rely on real-time adaptive changes to algorithmic parameters, such as \cite{Koide2021_adaptive}, or simplifying the algorithm to reduce the number of parameters or to try and remove the need for tuning altogether \cite{Vizzo2023_kiss_icp}. 

In this work we aim to decrease the tuning effort by increasing the understanding of the LIO algorithm parameters, narrowing the parameter search space, and facilitating the tuning process. The tuning effort will be decreased in two ways. First, by identifying parameters of high and low influence on performance, allowing the users to focus their efforts on parameters of high influence (i.e. reducing the number of dimensions of the parameter search space), and secondly, by identifying overarching trends in parameter influence based on the LiDAR characteristics, environment, flight characteristics, among others. The analysis is applied to two widely-used LiDAR-based LIO algorithms: Cartographer \cite{Hess2016} and \FL2 \cite{Xu2022_Fast_LIO2}. The methodology adopted is based on analyzing the effects of the algorithms' parameters on the quality of the estimated trajectories in UAV flight datasets in different types of scenarios and different LiDARs. To achieve this goal, we gather a large amount of LIO performance data resulting from executing Cartographer and FAST-LIO2 with different parameter choices, in diverse environments, with different LiDAR sensors and flight characteristics. The resulting data is then processed using Pearson's correlation and random forests to determine the parameters that have a strong influence on the LIO performance and their effects. The paper also provides tuning recommendations for Cartographer and \FL2 and a simplified tuning procedure consisting in manually tuning the two most influential parameters of each algorithm and leaving the others on the proposed recommended values. On the analyzed datasets, the proposed tuning procedure obtained a performance that is comparable to the performance of optimal parameter combinations found using exhaustive grid search.

To support the obtained results and contribute to deeper understanding of LIO tuning, all the processed data will be made publicly available as supplementary material upon paper acceptance. 

The main contributions of this paper are:

\begin{itemize}
    \item parameter sensitivity analysis for Cartographer and \FL2 on public datasets using Pearson's correlation and random forest permutation importance analysis,
    \item identification of parameters strongly influential in the performance for Cartographer and \FL2 algorithms,
    \item derivation of simplified parameter tuning recommendations and procedures for Cartographer and \FL2, and validation of such procedures and recommendations,
    \item over 68 GB / 6,900 hours of LIO estimated trajectory data and 140,000 LIO performance data are provided to contribute to deeper understanding of LIO tuning.
\end{itemize}

The rest of the paper is organized as follows. Section \ref{sec:related_work} presents previous work on SLAM and LIO parameter exploration. 
Section \ref{sec:methodology} presents the adopted methodology including the theoretical background and processing pipeline, and the choice of analyzed parameters for both LIO algorithms. Section \ref{sec:results} summarizes the results and findings obtained for each LIO algorithm including the identification of the parameters with the strongest influence, the tuning recommendations, and a simplified tuning procedure. Section \ref{sec:conclusion} concludes the paper and lays out guidelines for future work.

\section{Related work}
\label{sec:related_work}

Most previous work on  SLAM and LIO parameter exploration was conducted as part of a wider design space exploration consisting of parameters of the algorithms, of the hardware parameters, and of the compiler. Depending on the specific use case, the optimization criteria typically considered metrics of robot pose accuracy, algorithm runtime, and power consumption. For example, work \cite{Zia2016} explored the hardware parameters and algorithmic parameters for two visual SLAM methods. The hardware parameters consisted of choosing between a desktop computer and an ARM-based ODROID, and the number and type of CPU cores used, among others. Algorithmic parameters were explored using SLAM quality metric based on the absolute trajectory error (ATE), and the runtime and power consumption. Work \cite{Hernandez2021} applies Shapley values from game theory to measure the impact of algorithmic and platform-specific parameters on accuracy and power consumption on resource-constrained devices.

A design exploration framework is introduced by Bodin \cite{Bodin2016_PACT} to optimize the computational cost of the KinectFusion 3D scene understanding \cite{Newcombe_2011_KinectFusion} algorithm such it can run in real-time in low-power system on a chip (SoC) hardware. The work explores the influence of hardware, compiler, and algorithmic parameters on 3D reconstruction accuracy, runtime, and power consumption. The analysis was done only on one data sequence from the ICL-NUIM RGB-D Visual Odometry, 3D reconstruction and SLAM dataset \cite{Handa_ICRA2014}. The work is expanded in \cite{Nardi2017_PACT} to cover more diverse hardware architectures (such as desktop computers) and an additional visual SLAM algorithm, ElasticFusion \cite{Whelan2015_elastic_fusion}.
As the parameter space in \cite{Bodin2016_PACT} and \cite{Nardi2017_PACT} spans millions of combinations, the parameter space was randomly sampled for combinations which were tested for evaluation. The obtained performance data was used to train random forest regressors \cite{Breiman2001_random_forests} for each metric. These estimators were used to predict the performance over the entire parameter space (beyond the already sampled combinations) with additional parameter sets being sampled near the determined Pareto front. 
In addition to random forests, the authors explored a number of other base predictive models, including artificial neural networks, support vector machines, and nearest neighbors, but confirmed that random forests outperformed the other methods. However, no data is presented on the performance of these models or on the performance of the random forests used in the works.

In Saeedi \cite{Saeedi2017_ICRA} the parameter space is expanded with motion structure parameters and used it to apply movement constraints to a robotic arm and guide active visual SLAM.

Unlike the works \cite{Bodin2016_PACT}, \cite{Nardi2017_PACT}, and \cite{Saeedi2017_ICRA}, which used space exploration to find Pareto optimal combinations for RGB-D SLAM, our work finds general trends in LiDAR-based algorithm performance in order to reduce the tuning effort and provide parameter tuning recommendations. 
We use the random forest permutation-based feature importance method to analyze the magnitude of the parameters' influence on the performance metric. By this we expand the scope of the analysis to allow: i) the identification of highly influential parameters uncorrelated to the performance metric (which require manual tuning), ii) differentiation between high and low influence parameters, and iii) the proposal of a recommended tuning order. Random forest feature importances have long been used for exploratory data analysis and variable selection in other fields \cite{Ishwaran2007_var_importances_in_rfs, Genuer2010_variable_selection_rfs, Louppe2013_understanding_rf_importances}. 
To the best of our knowledge, this is the first time in which random forest permutation-based feature importance analyses have been applied to SLAM and LIO parameter analysis.

Our work provides parameter tuning recommendations for LiDAR-based LIO and SLAM algorithms for aerial robot autonomous navigation operations. To cope with the strong influence of the environment, LiDAR sensor, and flight conditions, our work analyzes parameter choice considering datasets with 60 different combinations of LiDAR, types of environment, and flights speed. 
In addition, our analysis is performed adopting exhaustive parameter search instead of sampling the parameter space, which increases the accuracy of the correlation analysis as it is conducted on real data instead of modeled data. This approach allows us to validate the random forest results by splitting the data into training and testing datasets. 
Unlike previous works which focus on low-power hardware platforms, we are interested in the accuracy of LIO algorithms on medium-large uncrewed quadrotors with sufficient onboard computational capacity where the computer architecture, compilers, runtime, and power consumption are of lower importance, focusing specifically on accuracy.

Our parameter analysis is applied to two widely-used algorithms which fall into different families of LiDAR-based SLAM and LIO based on their pose estimation method and map representation: Cartographer \cite{Hess2016} and \FL2 \cite{Xu2022_Fast_LIO2}. Cartographer is a graph-based SLAM with a bi-resolutional 3D grid map representation. It is one of the most widely-used LiDAR-based SLAM methods \cite{Petrlik2025_DARPA_SubT_Cartographer_on_UAV, Pak2022_Cartographer_on_UAV_ICCAS, Cahn_Cartographer_on_UAV_ICCAIS}, and is widely used by the authors of this manuscript
\cite{Milijas2021,Ivanovic2022,Markovic2022, Orsulic2021}. \FL2 \cite{Xu2022_Fast_LIO2} is an IKF-based (Iterated Kalman Filter) LIO algorithm with an \textit{ikd}-tree-based (incremental k-dimensional tree) map representation. FAST-LIO2 is well-known for its accuracy and has been quickly adopted by the community in many problems \cite{Chen2023_scirob_FAST_LIO2_on_UAV, Lu2025_TRO_FAST_LIO2_on_UAV, Xu_2025_RAL_FAST_LIO2_on_UAV, Tang_2023_IROS}. While Cartographer includes loop closure mechanisms, for comparison, these mechanisms were disabled in this work to only consider its graph-based LIO module.

\section{Methodology}
\label{sec:methodology}

\begin{figure}[pos=ht!]
    \centering
    \includegraphics[width=0.8\linewidth, trim={0cm, 15.0cm, 11.5cm, 0.0cm}, clip]{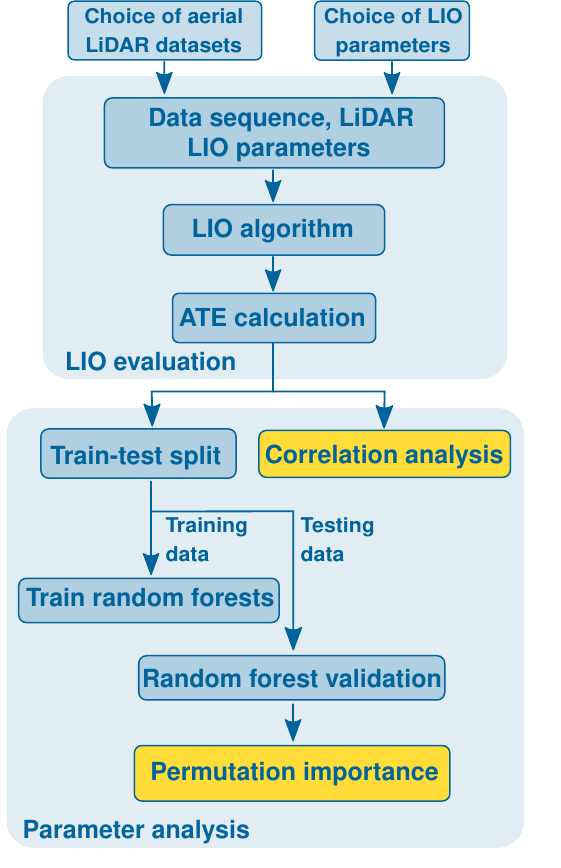}
    \caption{Summary of adopted methodology and the implemented processing pipeline. The final steps of the pipeline which were used for distilling tuning recommendations are highlighted with yellow color. 
    }
    \label{fig:methodology_summary}
\end{figure}

The goal of this work is to analyze the effects of different algorithmic parameters on the estimated trajectory quality with different LiDARs in different types of scenarios and flights, and to distill tuning recommendations for two widely-used LIO algorithms: Cartographer and FAST-LIO2. To achieve this goal, we gather and analyze a large amount of LIO performance data resulting from executing Cartographer and FAST-LIO2 with different parameter choices, in diverse environments, with different LiDAR sensors and flight characteristics. The gathered data is then analyzed to determine the parameters that have a strong influence on the LIO performance and their effects, as well as the overarching trends to aid parameter tuning. Figure \ref{fig:methodology_summary} shows the methodology adopted in the presented analysis, which can be broken down into two major parts: i) \textbf{LIO evaluation}, which chooses and evaluates the different LIO parameter combinations on different data sequences, and ii) \textbf{Parameter analysis}, which performs the analysis of the influence of the values of algorithms' parameters on the accuracy performance metric.

The analysis is conducted on high-quality data recorded from multirotor UAVs performing different types of flights at low-to-moderate altitudes and with close-to-moderate distances from obstacles in different environments (the conditions in which SLAM and LIO methods are widely used in many autonomous UAV navigation missions).
We choose two datasets (NTU VIRAL dataset \cite{Nguyen_2022_NTU_dataset} and PASTEL dataset \cite{Milijas2025}). In particular, PASTEL sequences were recorded in 16 flights in different environments with 3 onboard LiDAR sensors, enabling also to analyze the influence of the parameters with different LiDAR sensors.
The analysis was conducted on every combination of data sequence and LiDAR available in these datasets (60 in total) to explore the influence of LiDAR model and flight's and environment's characteristics on LIO tuning. The choice of datasets is discussed in Section \ref{subsec:aerial_slam_datasets}.

To obtain the LIO performance data required for the analysis, a grid search of the algorithms' parameter space was conducted. 
As the parameter spaces of the analyzed algorithms are very large 
(several tens of thousands of parameters combinations for Cartographer and for \FL2), to reduce the search space, some parameters of interest were chosen based on the  theoretical knowledge of their effect in the algorithms and single parameter tests, see Sections \ref{subsec:carto_params_chosen} and \ref{subsec:fastlio_params_chosen}. 
In \textbf{LIO evaluation}, shown at the top of Figure \ref{fig:methodology_summary}, for each of the chosen parameters, values corresponding to low, medium, and high settings were chosen to form the parameter grid search. The chosen parameter values and the LIO evaluation pipeline are laid out in Section \ref{subsec:processing_pipeline}. 
The absolute trajectory error (ATE), which is widely used to evaluate LIO accuracy, was calculated for each combination of parameters. The obtained data, called \emph{LIO performance data}, is the starting point of the parameter analysis.

\textbf{Parameter analysis}, shown at the bottom of Figure \ref{fig:methodology_summary}, performs the analysis of the \emph{LIO performance data} to extract the effects of the algorithms' parameters on performance. We use Pearson's correlation to determine the relationships between the algorithms' parameters and the algorithms' accuracy (ATE), see Section \ref{subsec:methodology_correlation_analysis}. 
A strong positive correlation indicates that an increase in the parameter value leads to an increase in the ATE, making smaller values favorable. Likewise, strong negative correlations indicate better performance with large values of the parameter. However, weak correlations indicate that the algorithm is either insensitive to the parameter or that the relationship between the parameter and the ATE metric is ambiguous.  

Additionally, the magnitude of a parameter's influence on the ATE allows differentiating between ambiguous cases which require manual tuning, and cases of weak parameter influence. Those cases are distinguished by analyzing the relative importance of parameters.
To analyze the magnitude of a parameter's influence we employ random forest regressors trained to predict the ATE metric based on the choice of LIO parameters. While random forests have been used to model the influence of parameters on SLAM performance, in our work we expand on this by utilizing 
permutation-based feature importance analysis to quantify the magnitude of the parameters' influence on the random forest models' predictions and to analyze the parameters' relative importance, see Section \ref{subsec:random_forests}. 
To ensure that the data obtained from the random forest models is applicable to the analyzed LIO algorithms, the models were tested on unseen data. This was achieved by splitting the obtained \emph{LIO performance data} between training (80\%) and testing (20\%) data. 
The widely-used Python library Scikit-learn \cite{scikit-learn} was used to implement the random forests and to calculate the permutation-based feature importances.

Finally, the analyses are distilled into tuning recommendations, which specify the order in tuning, recommended parameter values, and policies. These tuning recommendations are validated on both algorithms obtaining on the analyzed datasets ATE values within 5 cm to the optimal performance found in the grid search procedure in 94\% of the analyzed cases.

\subsection{Overview of used aerial SLAM datasets}
\label{subsec:aerial_slam_datasets}

The LIO and SLAM community has long relied on high-quality public datasets which facilitate testing in diverse scenarios, ensuring repeatability and transparency. 
In our research we are not interested in aerial LiDAR datasets captured at high altitudes, as the distances to the obstacles have low variability along the flight, and in general don't reproduce the conditions of LiDAR-based SLAM and LIO operation in typical autonomous UAV navigation missions. Instead, we choose datasets
recorded from multirotor UAVs performing different types of flights at low-to-moderate altitudes and with close-to-moderate distances from obstacles in different environments. 
Datasets with these characteristics provide rich and diverse tuning scenarios for LIO algorithms to ensure that we excite as much parameter dynamics 
as possible.

\begin{table*}[pos=ht!]
\caption{Data sequences used in this research with characteristics and statistics relevant for LIO performance: environment, average robot velocity, traversed distance, covered area, time, average and maximum altitude, and percentage of LiDAR points with range lower than 5, 10, and 20 m. The table also links the original sequence names with the IDs assigned in this research.
}
\label{table:combined_dataset}
\scalebox{0.9}{
\begin{tabular}{lclllllllll}
\toprule
\multicolumn{11}{c}{\textbf{PASTEL \cite{Milijas2025} sequences used as input data}} \\
\midrule
\multirow{2}{1.4cm}{\textbf{Sequence file name}} &
\mrII{\textbf{ID}} &
\mrII{\textbf{Environment}} &
\multirow{2}{1.25cm}{\textbf{Average velocity}} &
\multirow{2}{1.25cm}{\textbf{Traversed distance}} &
\multirow{2}{0.8cm}{\textbf{Covered area }} &
\multirow{2}{*}{\textbf{Time }} &
\multirow{2}{1.25cm}{\textbf{Max. altitude}} &
\multicolumn{3}{c}{\textbf{LiDAR points with range}} \\
&&&&&&&& \textbf{\textless \ 5m} & \textbf{\textless \ 10 m} & \textbf{\textless \ 20 m }\\
\midrule
1.bag   & 1  & EMPTY PLAZA         & 0.91 m/s & 61  m & 130  m$^2$ & 87  s & 2.75 m  & 12.2\%  & 33.7\%  & 63.8\%  \\
2.bag   & 2  & EMPTY PLAZA         & 0.43 m/s & 31  m & 78   m$^2$ & 97  s & 2.14 m  & 30.5\%  & 61.5\%  & 79.3\%  \\
3.bag   & 3  & SMALL CORRAL        & 0.72 m/s & 41  m & 16   m$^2$ & 97  s & 1.10 m  & 69.2\%  & 95.9\%  & 97.9\%  \\
4.bag   & 4  & SMALL CORRAL        & 0.56 m/s & 43  m & 17   m$^2$ & 104 s & 1.44 m  & 73.5\%  & 98.2\%  & 99.1\%  \\
5.bag   & 5  & LARGE CORRAL        & 0.69 m/s & 26  m & 15   m$^2$ & 71  s & 1.19 m  & 64.2\%  & 91.7\%  & 96.4\%  \\
6.bag   & 6  & LARGE CORRAL        & 0.53 m/s & 23  m & 20   m$^2$ & 63  s & 0.5  m  & 67.5\%  & 93.7\%  & 97.6\%  \\
7.bag   & 7  & UNDERPASS           & 0.98 m/s & 57  m & 119  m$^2$ & 79  s & 1.25 m  & 24.9\%  & 62.5\%  & 95.1\%  \\
8.bag   & 8  & UNDERPASS           & 0.72 m/s & 52  m & 103  m$^2$ & 91  s & 1.18 m  & 24.5\%  & 65.8\%  & 95.0\%  \\
9.bag   & 9  & PLAZA \& CORRAL     & 0.71 m/s & 37  m & 99   m$^2$ & 71  s & 4.84 m  & 29.9\%  & 51.1\%  & 68.2\%  \\
10.bag  & 10 & PLAZA \& CORRAL     & 0.63 m/s & 46  m & 70   m$^2$ & 88  s & 3.46 m  & 26.0\%  & 47.7\%  & 72.3\%  \\
11.bag  & 11 & PLAZA \& CORRAL     & 1.07 m/s & 155 m & 639  m$^2$ & 128 s & 1.46 m  & 21.1\%  & 50.2\%  & 72.8\%  \\
12.bag  & 12 & PLAZA \& CORRAL     & 0.95 m/s & 201 m & 1305 m$^2$ & 220 s & 1.51 m  & 22.1\%  & 56.3\%  & 78.2\%  \\
13.bag  & 13 & PLAZA \& UNDERPASS  & 1.31 m/s & 108 m & 421  m$^2$ & 81  s & 1.31 m  & 26.1\%  & 49.3\%  & 75.6\%  \\
14.bag  & 14 & PLAZA \& UNDERPASS  & 0.85 m/s & 182 m & 1365 m$^2$ & 261 s & 1.78 m  & 20.9\%  & 52.1\%  & 77.5\%  \\
15.bag  & 15 & ALL                 & 1.35 m/s & 253 m & 1819 m$^2$ & 186 s & 1.43 m  & 22.8\%  & 57.2\%  & 78.5\%  \\
16.bag  & 16 & ALL                 & 0.93 m/s & 138 m & 626  m$^2$ & 159 s & 5.54 m  & 19.7\%  & 40.6\%  & 69.9\%  \\
\midrule
\multicolumn{11}{c}{\textbf{NTU VIRAL dataset \cite{Nguyen_2022_NTU_dataset} }} \\
\midrule
Eee\_01.bag & 1  & Parking                      & 0.74 m/s & 237 m & 334 m$^2$ & 398 s & 9.97  m & 12.31\% & 20.0\% & 39.1\% \\
Eee\_02.bag & 2  & Parking                      & 0.68 m/s & 171 m & 284 m$^2$ & 321 s & 7.48  m & 13.50\% & 21.1\% & 49.4\% \\
Eee\_03.bag & 3  & Parking                      & 0.89 m/s & 128 m & 290 m$^2$ & 181 s & 4.75  m & 14.85\% & 24.6\% & 52.3\% \\
Nya\_01.bag & 4  & Large hall                   & 0.52 m/s & 160 m & 185 m$^2$ & 394 s & 6.91  m & 12.34\% & 28.6\% & 74.6\% \\
Nya\_02.bag & 5  & Large hall                   & 0.71 m/s & 249 m & 109 m$^2$ & 428 s & 5.39  m & 12.10\% & 25.9\% & 73.6\% \\
Nya\_03.bag & 6  & Large hall                   & 0.88 m/s & 315 m & 95  m$^2$ & 409 s & 6.72  m & 9.38\%  & 24.4\% & 73.9\% \\
Sbu\_01.bag & 7  & Small square                 & 0.72 m/s & 202 m & 291 m$^2$ & 354 s & 4.76  m & 14.33\% & 21.0\% & 59.5\% \\
Sbu\_02.bag & 8  & Small square                 & 0.60 m/s & 183 m & 253 m$^2$ & 373 s & 6.07  m & 12.02\% & 19.6\% & 62.8\% \\
Sbu\_03.bag & 9  & Small square                 & 0.63 m/s & 198 m & 362 m$^2$ & 389 s & 6.33  m & 11.04\% & 20.4\% & 59.5\% \\
Tnp\_01.bag & 10 & Inside res. tech. plaza & 0.54 m/s & 252 m & 226 m$^2$ & 579 s & 9.16  m & 9.10\%  & 27.6\% & 86.1\% \\
Tnp\_02.bag & 11 & Inside res. tech. plaza & 0.39 m/s & 140 m & 242 m$^2$ & 457 s & 10.99 m & 11.48\% & 33.5\% & 85.3\% \\
Tnp\_03.bag & 12 & Inside res. tech. plaza & 0.47 m/s & 124 m & 224 m$^2$ & 407 s & 9.82  m & 21.07\% & 34.8\% & 87.8\% \\
\bottomrule
\end{tabular}
}
\end{table*}

We apply our research to the combination of PASTEL \cite{Milijas2025} and NTU VIRAL \cite{Nguyen_2022_NTU_dataset} datasets, which together provide high diversity in the main factors affecting LIO tuning: LiDAR characteristics, environment types (confined spaces, open spaces of different sizes, and transitions between them), LiDAR point distributions, flight velocities, and flight altitudes, among others. Table \ref{table:combined_dataset} shows the most relevant characteristics of the sequences used in this research. The table links the original file names of the datasets with the IDs assigned to them for our research, and lists the environment names taken from the both publications. The distance traversed by the UAV and area covered by the trajectory for each sequence was calculated from the provided ground truth. The NTU VIRAL dataset was recorded on higher average altitudes and with larger distances from obstacles than PASTEL, which makes them complementary. The table also shows the percentage of LiDAR points with a range lower than 5 m, 10 m, and 20 m for each sequence, evidencing a wide variety of LiDAR point range distributions.

In addition, PASTEL provides scans from three different LiDAR models mounted horizontally and recorded in parallel: Velodyne VLP16, Ouster OS1-16, and Ouster OS0-128, allowing us to test the influence of different LiDARs on parameter tuning. The NTU VIRAL dataset uses two LiDARs, one mounted horizontally, and one mounted vertically to scan the floor. To keep the dataset performance comparable, only the horizontal LiDAR from NTU VIRAL was used. The model and revision of the NTU VIRAL LiDAR is identical to the Ouster OS1-16 LiDAR used in PASTEL.
The detailed characteristics of the sensors, flights, environments of these sequences are available in \cite{Milijas2025} and \cite{Nguyen_2022_NTU_dataset}.

\subsection{Cartographer parameters chosen for tuning}
\label{subsec:carto_params_chosen}

Being a graph-based SLAM with a bi-resolutional 3D grid representation, Cartographer is a complex system with many modules and tunable parameters. Since this work focuses on LIO algorithms, the analysis is constrained to the parameters of the Cartographer LIO subsystem (called local SLAM in the original paper \cite{Hess2016}), while the loop closing subsystem (called global SLAM) is disabled.
Information on all the Cartographer subsystems and descriptions of parameters can be found online\footnote{\url{https://google-cartographer-ros.readthedocs.io}, accessed on Jan 9th 2026}. 

The Cartographer LIO subsystem contains several submodules with +30 tunable parameters. We narrow the search space to exclude parameters which 
either depend heavily on the used hardware (e.g. number of LiDAR data packets per full sweep, minimum LiDAR range, and number of CPU cores allocated to the solver), or perform additional internal data filtering beyond the selected parameters (e.g. minimal motion requirement to keep a LiDAR scan). 
The selected parameters and their main effects are listed in Table \ref{table:cartographer_parameter_summary}.
The non-considered parameters were left in their default values. We evaluated each data sequence and LiDAR combination with these default values and ensured that they do not cause coarse performance degradations, such as sparse or empty submaps, and that the LiDAR minimum range parameter crops only the UAV frame from the sweeps without cropping parts of the obstacles. These tests also confirmed that these default values are adequate for Cartographer to run faster than real time for all considered data sequences and sensors.

\begin{table}[pos=ht!]
    \caption{Summary of the Cartographer parameters selected for analysis.
    }
    \label{table:cartographer_parameter_summary}
    \centering
    \begin{tabular}{l@{\hspace{1mm}}|l}
    \toprule
    \multicolumn{2}{c}{\textbf{Selected Cartographer parameters and main influences
    }} \\
    \midrule
      \multirow{3}{*}{\textbf{Rotation\_weight}}
            & \multirow{6}{4.8cm}{Weight of IMU preintegration when determining the pose estimate. Higher value causes smoother orientation and translation estimates.} \\ \\  \\
      \cline{1-1}
      \multirow{3}{*}{\textbf{Translation\_weight}} & \\ \\ \\
      \midrule
      \multirow{3}{*}{\textbf{Low\_resolution}}   
            & \multirow{6}{4.8cm}{Grid resolutions for the bi-resolutional submaps. Coarser maps increase computation speed and reduce memory requirements but reduce level of captured detail and vice-versa.} \\ \\ \\
      \cline{1-1}
      \multirow{3.5}{*}{\textbf{High\_resolution}}  
            & \\ \\ \\
            \midrule
      \multirow{4}{*}{\textbf{Submap\_size}}  
            & \multirow{4}{4.8cm}{Number of LiDAR scans per submap. Larger submaps hold more information for scan matching but also accumulate more drift.} \\ \\ \\ \\
            \midrule
      \multirow{3}{*}{\textbf{Voxel\_filter\_size}} 
            & \multirow{3}{4.8cm}{Reduces input LiDAR scan density. Denser scans hold more detail but require more processing power.} \\ \\ \\
            \midrule
      \multirow{3}{*}{\textbf{LiDAR\_max\_range}}
            & \multirow{3}{4.8cm}{Long range LiDAR data can be noisy, sparse and contaminated with LiDAR beam reflections.} \\ \\ \\
        \bottomrule
    \end{tabular}
\end{table}

\subsection{\FL2 parameters chosen for tuning}
\label{subsec:fastlio_params_chosen}

Analogously, the parameters that depend heavily on the used hardware such as the number of CPU cores used, and LiDAR minimum range were excluded from the analysis and left on their default values. 
We evaluated each data sequence and LiDAR combination with these default parameters to ensure that no coarse performance degradations occur, and that the algorithm can run faster than real time. This narrowed the parameter space to ten parameters. 
To further narrow the choice of parameters a coarse sensitivity test was performed on PASTEL sequences. One by one, the parameters were changed from their default values by multiplying the default by 10 and by 0.1. The algorithm performance was evaluated on the estimated trajectory and mapping result.
All parameters which did not produce a strong change in SLAM performance were discarded from the analysis.
In addition, although parameter \textbf{LiDAR\_max\_range} is not originally contained in FAST-LIO2, a simple prefiltering step was added into the pipeline. 
Filtering LiDAR points by a range cutoff is a common preprocessing step in many LIO algorithms and strongly influences performance in many methods including Cartographer. In fact, as shown in Section \ref{sec:results}, this parameter showed to have the strongest influence in Cartographer. For homogenization, we decided to add this preprocessing step to investigate whether it has an influence on FAST-LIO2 as well. The prefilter was implemented in C++ as a ROS node which changes coordinates for points which exceed the range threshold to $(0, 0, 0)$. This is a standard procedure which many LiDAR drivers, such as Ouster ROS drivers, use to handle points with no returns, as keeping all the points preserves the organized point cloud structure. The parameters chosen for \FL2 are shown in Table \ref{table:fastlio2_parameter_summary}.

\begin{table}[pos=ht!]
    \caption{Summary of \FL2 parameters selected for analysis.
    }
    \label{table:fastlio2_parameter_summary}
    \centering
    \begin{tabular}{l|l}
    \toprule 
    \multicolumn{2}{c}{\textbf{Selected \FL2 parameters and main influence}} \\
    \midrule
      \mrII{\textbf{Acceleration\_covariance}}
            & \multirow{3}{4.0cm}{Kalman filter covariances for IMU accelerations and angular velocities.} \\  \\
      \cmidrule{1-1}
      \textbf{Gyroscope\_covariance}
            & \\
            \midrule
      \multirow{2}{*}{\textbf{Point\_filter\_number}}   
            & \multirow{2}{4.0cm}{LiDAR points prefiltering. Keeps every $N^{th}$ point.} \\ \\
            \midrule
      \multirow{3}{*}{\textbf{Filte\_size\_surfaces}}  
            & \multirow{3}{4.0cm}{Additional LiDAR prefilter. The points in each voxel are replaced by their centroid.} \\ \\ \\
            \midrule
      \multirow{3}{*}{\textbf{Filter\_size\_map}} 
            & \multirow{3}{4.0cm}{Voxel-based map downsampling. Analogous to filter size surfaces.} \\ \\ \\
            \midrule
      \multirow{4}{*}{\textbf{LiDAR\_max\_range}}
            & \multirow{4}{4.0cm}{Range-based LiDAR prefilter added in this work for comparison with Cartographer.} \\ \\ \\ \\
            \bottomrule
    \end{tabular}
\end{table}

\subsection{LIO evaluation pipeline}
\label{subsec:processing_pipeline}

To research the influence of the chosen parameters on LIO performance, an automated framework for evaluating the algorithms for a given data sequence, LiDAR sensor, and parameter set was developed. The parameter values for Cartographer and FAST-LIO2 chosen to form the parameter grid searches are reported in Table \ref{table:grid_search}. The Cartographer parameter values were chosen from our long experience in using Cartographer in a broad variety of scenarios and LiDARs (we call them LARICS parameters), while the \FL2 parameters were obtained by doubling and halving the default values provided by \FL2 authors in \cite{Xu2022_Fast_LIO2}.

\begin{table}[ht!]
    \centering
    \caption{Parameter values used in the 
    grid search. The LARICS parameters for Cartographer and default parameters for FAST-LIO2 are marked in bold. 
    }
    \label{table:grid_search}
    \begin{tabular}{lc}
    \toprule
    \multicolumn{2}{c}{\textbf{Choice of Cartographer parameter values}} \\
    \midrule
        \textbf{Rotation\_weight}              & {\textbf{5}, 10, 100}  \\
        \textbf{Translation\_weight}           & {\textbf{5}, 10} \\ 
        \textbf{High\_resolution}              & {5 cm, 10 cm, \textbf{20 cm}} \\
        \textbf{Low\_resolution}               & {30 cm, \textbf{40 cm}, 50 cm} \\
        \textbf{Submap\_size} (number of scans) & {150, \textbf{500}, 1000} \\ 
        \textbf{Voxel\_ filter\_size}           & {5 cm, \textbf{10 cm}, 15 cm} \\
        \textbf{LiDAR\_Max\_range}              & {20 m, 35 m, \textbf{50 m}} \\
    \midrule
    \multicolumn{2}{c}{\textbf{Choice of FAST-LIO2 parameter values}} \\
    \midrule
        \textbf{Acceleration\_covariance       } & 0.05, \textbf{0.1}, 0.2  \\
        \textbf{Gyroscope\_covariance          } & 0.05, \textbf{0.1}, 0.2 \\
        \textbf{Point\_filter\_number          } & 1, 2, \textbf{4}, 8 \\
        \textbf{Filter\_size\_surfaces         } & 25 cm, \textbf{50 cm}, 99 cm \\
        \textbf{Filter\_size\_map              } & 25 cm, \textbf{50 cm}, 99 cm \\
        \multirow{2}{3cm}{\textbf{LiDAR\_max\_range} (added as prefilter)} & \mrII{20 m, 35 m, 50 m } \\ \\
    \bottomrule
    \end{tabular}
\end{table}

The evaluation pipeline is shown in Figure \ref{fig:methodology_summary}. The LIO algorithms were executed for each combination of data sequence, LiDAR, and LIO parameter choice from the grid search. Each resulting robot trajectory estimate was recorded as it would be fed into a UAV control system. This evaluation resulted in over 140,000 estimated trajectories. They were computed using two identical ASUS NUC 13 computers, one per each LIO algorithm, running Ubuntu 20.04 and ROS Noetic. The estimated robot trajectories are extracted into \textit{.csv} files compatible with the NTU VIRAL open-source MATLAB-based SLAM evaluation framework \cite{Nguyen_2022_NTU_dataset}, which was used to compute the ATE metric (\ref{eqn:ATE}) for each estimated trajectory.

\begin{equation}
\label{eqn:ATE}
    \text{ATE} = \sqrt{
    \frac{1}{N} \sum^{N-1}_{i=0}|| \Delta \mathbf{p}_i ||^2
    }
\end{equation}

To validate that the chosen parameter space contains parameters with good LIO performance, Table \ref{table:best_grid_search_ATEs} shows the best ATE metrics found by the grid search for each algorithm, data sequence, and LiDAR combination. It can be noticed that the ATE values obtained for PASTEL sequences are lower than those obtained by Cartographer and by FAST-LIO2 using their default values, which can be found in \cite{Milijas2025}. On average, the decrease in ATE is 14 cm (54\%), evidencing the importance of correctly choosing the parameter values and the strong improvement they can cause. The results obtained for NTU VIRAL sequences cannot be compared, as they were validated with LIO-SAM \cite{Shan2020_LIOSAM} and MLOAM (multi-lidar LOAM), but not with Cartographer or \FL2, see \cite{Nguyen_2022_NTU_dataset}. As shown, the ATE values obtained for all sequences are very low, and validate the selection of the values used in the grid search.

\begin{table}[pos=ht!]
\centering
\caption{
Best ATE metrics found by the grid search for each algorithm, data sequence, and LiDAR combination. The ATE values obtained for all sequences are very low and validate the selection of the values used in the grid search.
}
\label{table:best_grid_search_ATEs}
\scalebox{0.90}{
\begin{tabular}{@{}l@{}c@{\hspace{1mm}}c@{\hspace{1mm}}c@{\hspace{1mm}}|c@{\hspace{1mm}}c@{\hspace{1mm}}c@{}}
\toprule 
\multicolumn{7}{c}{\textbf{Best ATE values [m] obtained in grid search}} \\
\midrule
&\multicolumn{3}{c}{Cartographer} & \multicolumn{3}{c}{FAST-LIO2} \\
\midrule
Data & \mrII{VLP16} & \mrII{OS1-16}   & \mrII{OS0-128}    & \mrII{VLP16} & \mrII{OS1-16}   & \mrII{OS0-128} \\ 
sequence   &       &          &            &       &          &          \\ 
\midrule                                             
PASTEL 1   & 0.100 & 0.138 & 0.090  & 0.064 & 0.014 & 0.063\\
PASTEL 2   & 0.036 & 0.056 & 0.053  & 0.075 & 0.103 & 0.069\\
PASTEL 3   & 0.045 & 0.077 & 0.057  & 0.056 & 0.079 & 0.061\\
PASTEL 4   & 0.024 & 0.071 & 0.063  & 0.071 & 0.094 & 0.074\\
PASTEL 5   & 0.049 & 0.025 & 0.065  & 0.048 & 0.072 & 0.063\\
PASTEL 6   & 0.044 & 0.099 & 0.077  & 0.046 & 0.069 & 0.042\\
PASTEL 7   & 0.066 & 0.128 & 0.118  & 0.055 & 0.098 & 0.077\\
PASTEL 8   & 0.057 & 0.100 & 0.080  & 0.053 & 0.080 & 0.037\\
PASTEL 9   & 0.050 & 0.088 & 0.056  & 0.061 & 0.081 & 0.054\\
PASTEL 10  & 0.057 & 0.101 & 0.095  & 0.061 & 0.081 & 0.055\\
PASTEL 11  & 0.008 & 0.162 & 0.028  & 0.082 & 0.163 & 0.140\\
PASTEL 12  & 0.087 & 0.165 & 0.131  & 0.104 & 0.183 & 0.128\\
PASTEL 13  & 0.105 & 0.125 & 0.184  & 0.079 & 0.165 & 0.150\\
PASTEL 14  & 0.080 & 0.135 & 0.118  & 0.099 & 0.166 & 0.134\\
PASTEL 15  & 0.118 & 0.240 & 0.187  & 0.095 & 0.185 & 0.136\\
PASTEL 16  & 0.082 & 0.155 & 0.170  & 0.064 & 0.130 & 0.073\\
\midrule
NTU VIRAL 1  & & 0.211 & & & 0.203  & \\
NTU VIRAL 2  & & 0.221 & & & 0.187  & \\
NTU VIRAL 3  & & 0.177 & & & 0.207  & \\
NTU VIRAL 4  & & 0.214 & & & 0.205  & \\
NTU VIRAL 5  & & 0.186 & & & 0.204  & \\
NTU VIRAL 6  & & 0.220 & & & 0.195  & \\
NTU VIRAL 7  & & 0.215 & & & 0.226  & \\
NTU VIRAL 8  & & 0.220 & & & 0.223  & \\
NTU VIRAL 9  & & 0.203 & & & 0.215  & \\
NTU VIRAL 10 & & 0.219 & & & 0.190  & \\
NTU VIRAL 11 & & 0.365 & & & 0.075  & \\
NTU VIRAL 12 & & 0.234 & & & 0.196  & \\
\bottomrule
\end{tabular}
} 
\end{table}

\subsection{Correlation analysis}
\label{subsec:methodology_correlation_analysis}

Pearson's correlation (\ref{eqn:PearsonR}) measures the strength of a linear relationship between two variables. It is often used for analyzing the effect of an input variable on a response variable. The correlation $r$ between an input $x$ and a response $y$ is calculated as follows:
\begin{equation}
    r = \frac{\sum_{i=1}^{n}(x_i - \bar{x})(y_i - \bar{y})}{\sqrt{\sum_{i=1}^{n}(x_i - \bar{x})^2}\sqrt{\sum_{i=1}^{n}(y_i - \bar{y})^2}},
    \label{eqn:PearsonR}
\end{equation}
\noindent
where $\bar{x}$ and $\bar{y}$ indicate the sample means of the input and response variables, respectively.

We use the Pearson's correlation coefficient to determine  the relationships between the chosen parameter and the resulting ATE. A positive correlation indicates that choosing a higher value of a parameter is expected to lead to an increase of the ATE value, while negative correlations indicate lower ATE for larger values of the parameter. However, correlation values alone are insufficient for distilling tuning recommendations because weak correlations (values close to zero) can indicate that the parameter has a weak or ambiguous influence on ATE. 

\begin{figure*}[pos=ht!]
    \centering
     \includegraphics[width=0.99\linewidth, trim={0.75cm, 7.15cm, 0.5cm, 7.65cm}, clip]{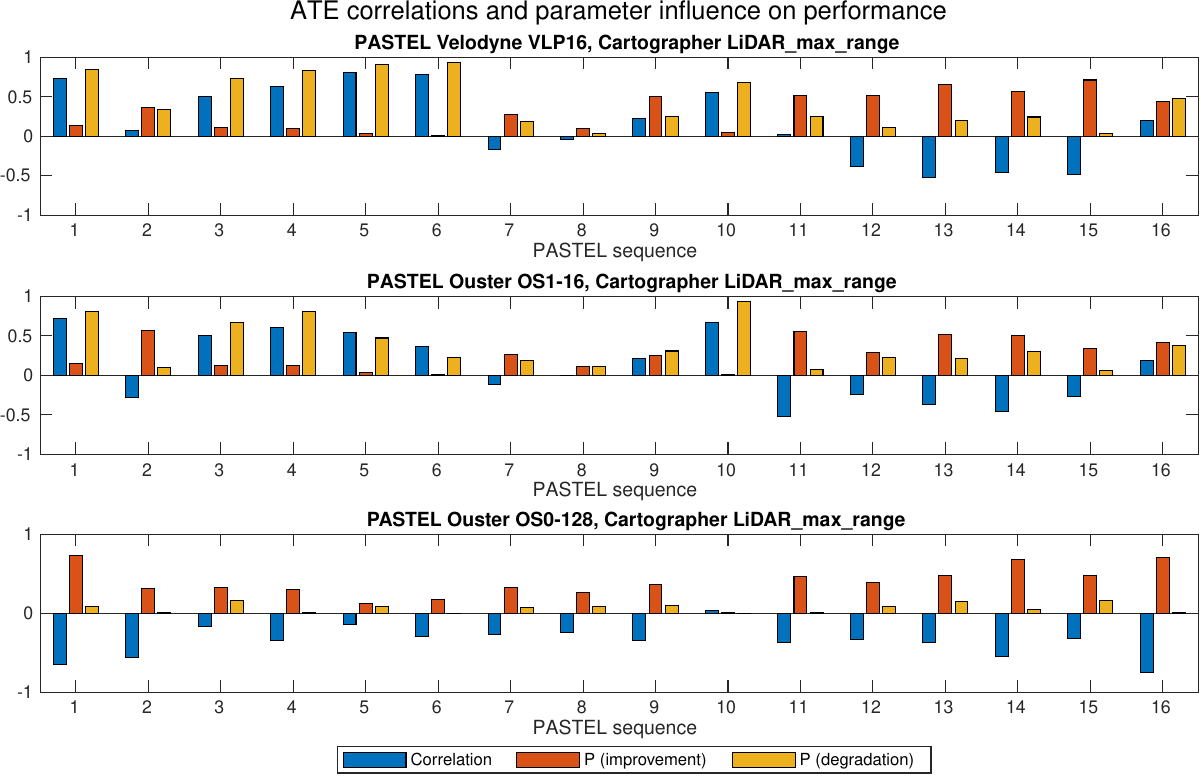}
     \caption{Comparison of the correlation between ATE and \textbf{LiDAR\_max\_range} of Cartographer performance on the sequences of PASTEL dataset. The bars indicate: i) correlation of \textbf{LiDAR\_max\_range} with ATE (blue color), ii) probability of improving the ATE by more than 1 cm when increasing \textbf{LiDAR\_max\_range} (red); and iii) probability of degrading the ATE by more than 1 cm when increasing \textbf{LiDAR\_max\_range} (yellow).}
     \label{subfig:cartographer_correlations_validation}
\end{figure*}

Figure \ref{subfig:cartographer_correlations_validation}
illustrates these effects on \emph{LIO performance data} for Cartographer parameter \textbf{LiDAR\_max\_range} on all of the 16 sequences in PASTEL dataset and with the 3 LiDAR sensors considered in PASTEL: Velodyne VLP16 (top), Ouster OS1-16 (center), and Ouster OS0-128 (bottom). The bar chart for each sequence and LiDAR displays the result of three analyses in different colors: i) correlation of \textbf{LiDAR\_max\_range} with ATE (blue color), ii) probability of improving the ATE by more than 1 cm when increasing \textbf{LiDAR\_max\_range} (red color); and iii) probability of degrading the ATE by more than 1 cm when increasing \textbf{LiDAR\_max\_range} (yellow color). The probabilities were calculated based on the \emph{LIO performance data}.  In some cases there is a strong positive correlation, indicating a high probability of LIO performance degradation when increasing \textbf{LiDAR\_max\_range} (e.g. for Velodyne VLP16 on sequences 3-6, which correspond to confined spaces as shown in Table \ref{table:combined_dataset}). Analogously, a strong negative correlation indicates a high probability of performance improvement (e.g. for Velodyne VLP16 on sequences 13-15). Weak correlation values can indicate ambiguous effects on LIO performance, (e.g. with Ouster OS1-16 and Velodyne VLP16 on sequence 16), or indifference to the parameter (e.g. all LiDARs in sequence 8). Similar effects were observed for all parameters in both algorithms on both datasets, which can be verified in the provided supplementary data.

Also, it must be noticed that the correlation coefficient holds no information about the magnitude of a parameter's influence on the ATE value: for a strongly correlated parameter, the ATE values may change in the order of centimeters or meters. Hence, correlations alone are not enough to determine which parameters to tune and in what order, since the information about the parameter relative importance and magnitude of influence on ATE are missing.
To analyze the magnitude of a parameter's influence, random forest regressors were trained to predict ATE values based on the choice of LIO parameters for a given data sequence and LiDAR. These trained models were used to extract information about the LIO parameters' influence using random forest feature importance methods.

\subsection{Random forest regressors and feature importance}
\label{subsec:random_forests}

\begin{figure}[pos=ht!]
    \centering
    \includegraphics[width=0.99\linewidth, trim={0cm, 18.0cm, 3.8cm, 0cm}, clip]{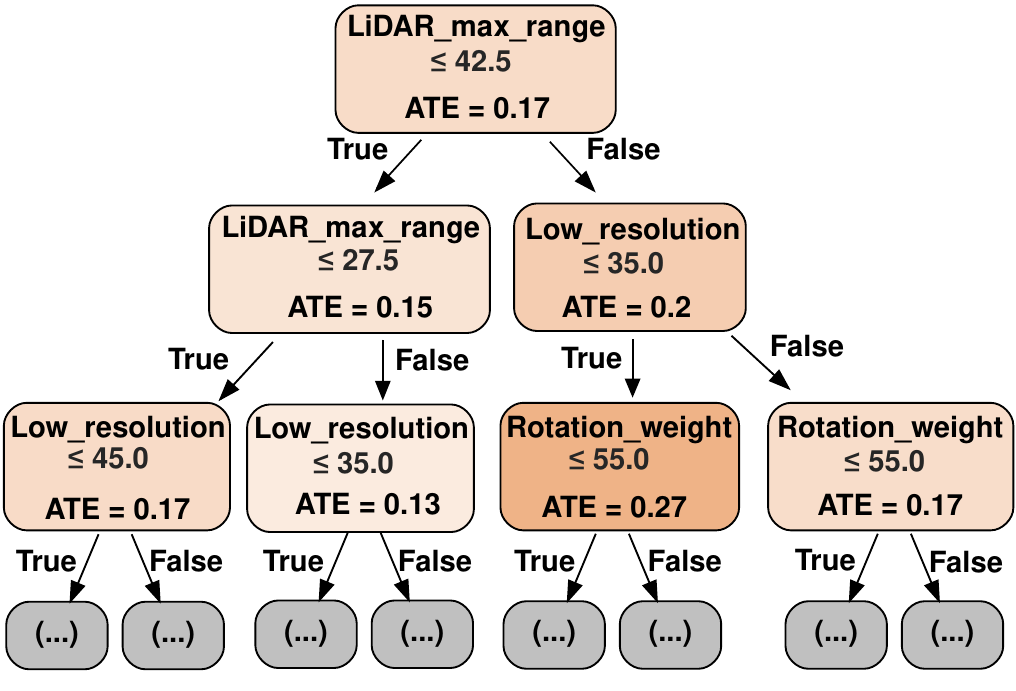}
    \caption{Scheme with the first three levels of a decision tree trained for estimating Cartographer performance on PASTEL sequence 16 with Velodyne VLP16 LiDAR. Each decision splits the parameter space using one of the parameters to form an if-else decision. 
    }
    \label{fig:decision_tree_example}
\end{figure}

Random forest regressors are machine learning ensemble models that are built from many individual decision trees which are trained on various random sub-samples of the input data to prevent overfitting and ensure randomness. The predictions of the individual decision trees are averaged to form the forest's final prediction \cite{Breiman2001_random_forests}. 
The underlying decision tree regressor uses the input features to split the approximated function's domain into sub-spaces of constant value. Each node in the tree represents an if-else decision based on the input feature which reduces the error criteria the most, while leaf nodes carry a predicted function value for their feature sub-space. 
We leveraged random forest regressors to model the influence of parameters on SLAM performance. The input features are the LIO parameter choices and the predicted function is the ATE metric. 
Figure \ref{fig:decision_tree_example} shows the first three levels of a decision tree trained on the \emph{LIO performance data} for Cartographer on PASTEL sequence 16 with the Velodyne VLP16 LiDAR.

\begin{table}[pos=ht!]

\caption{
Testing of trained random forest models for Cartographer. The RMSE scores obtained on the unseen testing data is compared with the standard deviations of ATEs in the testing data for each data sequence and LiDAR. The results are satisfactory with the predictors, obtaining a test RMSE which is on average 58\% lower than the standard deviation of the ATE testing data. 
}
\label{table:Cartographer_random_forest_validation}
\centering
\scalebox{0.9}{
\begin{tabular}{@{}ccc|ccc@{}}
\toprule
\multicolumn{3}{c}{\textbf{PASTEL, Velodyne VLP16}} & \multicolumn{3}{c}{\textbf{PASTEL, Ouster OS1-16}} \\
\mrII{Sequence}  & ATE      & Test                  & \mrII{Sequence}  & ATE      & Test   \\
                 & std dev  & RMSE                  &                  & std dev  & RMSE   \\
\midrule     
1                & 13.3 cm  & 3.5 cm                & 1                & 11.8 cm  & 2.5 cm \\
2                & 4.4 cm   & 2.1 cm                & 2                & 3.5 cm   & 1.3 cm \\
3                & 4.1 cm   & 1.3 cm                & 3                & 2.8 cm   & 1.5 cm \\
4                & 8.3 cm   & 3.5 cm                & 4                & 7.4 cm   & 3.6 cm \\
5                & 4.4 cm   & 1.3 cm                & 5                & 1.2 cm   & 0.8 cm \\
6                & 3.1 cm   & 1.1 cm                & 6                & 0.9 cm   & 0.3 cm \\
7                & 2.8 cm   & 0.9 cm                & 7                & 1.3 cm   & 1.0 cm \\
8                & 0.9 cm   & 0.3 cm                & 8                & 1.0 cm   & 0.5 cm \\
9                & 11.2 cm  & 4.7 cm                & 9                & 8.0 cm   & 6.0 cm \\
10               & 5.7 cm   & 1.9 cm                & 10               & 9.3 cm   & 2.8 cm \\
11               & 5.0 cm   & 2.0 cm                & 11               & 1.3 cm   & 0.5 cm \\
12               & 2.4 cm   & 1.0 cm                & 12               & 2.8 cm   & 0.8 cm \\
13               & 4.5 cm   & 1.5 cm                & 13               & 2.5 cm   & 1.7 cm \\
14               & 4.1 cm   & 1.2 cm                & 14               & 4.7 cm   & 1.2 cm \\
15               & 9.4 cm   & 4.3 cm                & 15               & 6.6 cm   & 4.0 cm \\
16               & 6.1 cm   & 2.9 cm                & 16               & 5.4 cm   & 2.4 cm \\
\midrule
\multicolumn{3}{c}{\textbf{PASTEL, Ouster OS0-128}} & \multicolumn{3}{c}{\textbf{NTU VIRAL, Ouster OS1-16}} \\
\mrII{Sequence}  & ATE      & Test                  & \mrII{Sequence}  & ATE      & Test   \\
                 & std dev  & RMSE                  &                  & std dev  & RMSE   \\
\midrule                                            
1                & 3.0 cm   & 1.7 cm                & 1                & 19.8 cm  &  4.5 cm \\
2                & 0.9 cm   & 0.4 cm                & 2                & 28.7 cm  & 13.0 cm \\
3                & 1.5 cm   & 1.0 cm                & 3                & 16.1 cm  &  2.1 cm \\
4                & 1.4 cm   & 0.5 cm                & 4                & 9.8 cm   &  2.4 cm \\
5                & 0.7 cm   & 0.6 cm                & 5                & 19.1 cm  &  4.5 cm \\
6                & 1.1 cm   & 0.3 cm                & 6                & 22.0 cm  &  6.9 cm \\
7                & 1.6 cm   & 0.8 cm                & 7                & 5.4 cm   &  1.1 cm \\
8                & 1.1 cm   & 0.7 cm                & 8                & 10.9 cm  &  3.0 cm \\
9                & 1.4 cm   & 1.3 cm                & 9                & 25.0 cm  &  6.4 cm \\
10               & 0.5 cm   & 0.2 cm                & 10               & 34.6 cm  & 12.3 cm \\
11               & 1.3 cm   & 1.0 cm                & 11               & 70.7 cm  & 17.8 cm \\
12               & 2.5 cm   & 0.9 cm                & 12               & 53.8 cm  & 14.2 cm \\
13               & 1.7 cm   & 1.3 cm                & /                & \multicolumn{2}{c}{/}       \\
14               & 4.5 cm   & 1.2 cm                & /                & \multicolumn{2}{c}{/}       \\
15               & 2.4 cm   & 1.4 cm                & /                & \multicolumn{2}{c}{/}       \\
16               & 3.9 cm   & 0.8 cm                & /                & \multicolumn{2}{c}{/}       \\
\bottomrule
\end{tabular}
} 
\end{table}

Conversely to existing methods, we expand the use of random forest regressors by 
making use of the permutation-based feature importance method to quantify the strength of the random forest input features' influence on the trained models' predictions. This enables quantifying the input features' relative importance and hence, the relevance of the input features used in the models. 
Permutation-based feature importance is calculated by permuting one of the features in the input data and validating the model on the permuted data. Since the model was trained on the non-permuted data, the permutation of a feature will decrease the model performance, breaking the link between that feature and the predicted value. The larger the drop in performance, the more sensitive the model is to that particular feature. We refer the reader to \cite{Genuer2010_variable_selection_rfs} for a theoretical background and discussion on permutation-based feature importance.

To ensure that the conclusions drawn from the models and analyses apply to the studied LIO algorithms, the trained random forest models were tested on unseen data points. This was achieved by training the random forests on 80\% of the input data (\emph{LIO performance data}) and using the remaining 20\% for testing. 
The models were tested by calculating the RMSE of the model predictions on the testing data. For example, the resulting values for random forests modeling Cartographer performance for each data sequence and LiDAR in the PASTEL and NTU VIRAL datasets are shown in Table \ref{table:Cartographer_random_forest_validation}. 
The table compares the model RMSE values obtained on the testing data with the ATE standard deviations in the testing data for each data sequence-LiDAR combination processed by Cartographer. 
The table shows that all predictors obtained satisfactory results with their test RMSEs being on average 58\% lower than the standard deviation of the testing ATE values, demonstrating that random forests can be used to predict SLAM ATE values well. 
The low test RMSE values obtained for PASTEL sequences 7 and 8 are consistent with the low ATE standard deviations for these sequences.

We used the random forest regressors and permutation-based feature importance method from the widely-used Python library Scikit-learn \cite{scikit-learn}. 
The models were analyzed for overfitting by increasing the minimum number of leaf data points and by varying the number of trees in the random forests. We observed no significant difference with the results in Table \ref{table:Cartographer_random_forest_validation}, which discards overfitting, ensuring the results correctness. 

\section{Results}
\label{sec:results}

This section presents and interprets the results obtained in the aforementioned analyses separately for each LIO algorithm. 
The analyses results are interpreted in three steps. We start with the random forest feature importances to determine the parameters with high and low influence on the algorithm's performance for each data sequence and LiDAR combination in the datasets. The second step interprets the correlation analysis data to determine which of the influential parameters require manual tuning. The third step determines overarching trends of these influential parameters. The interpretation of the results concludes with parameter tuning recommendations for each LIO algorithm, which are finally validated. 

\subsection{Cartographer parameter analysis}
\label{subsec:cartographer_analysis}

The influence of the different parameters in Cartographer is analyzed with the results shown in Table \ref{table:cartographer_permutation_importance} (permutation-based random forest feature importance) and Table \ref{table:cartographer_correlations} (correlation analyses). The rows are organized into four groups according to the LiDAR sensor, and each group contains the values obtained by each Cartographer parameter. Additionally, in Table \ref{table:cartographer_permutation_importance}, each group starts with the random forest prediction error (RMSE) obtained without permutation for the given data sequence and LiDAR while the rest are the RMSE values resulting when permuting each Cartographer parameter.

\begin{table*}[ht!]
\centering
\caption{Random forest feature permutations analysis for Cartographer parameters for each data sequence and LiDAR in the considered datasets. Each group starts with the random forest prediction error (RMSE) obtained without permutation, while the rest are the RMSE values resulting when permuting each specific Cartographer parameter. The higher the increase in RMSE caused by permutation of feature values, the higher the influence of the parameter. Cases which increase the RMSE by less than 50\% of the RMSE without permutation or by less than 2cm are marked in gray. All values are reported in centimeters. 
}
\label{table:cartographer_permutation_importance}
\scalebox{0.92}{
\begin{tabular}{cl|c@{\hspace{2.5mm}}c@{\hspace{2.5mm}}c@{\hspace{2.5mm}}c@{\hspace{2.5mm}}c@{\hspace{2.5mm}}c@{\hspace{2.5mm}}c@{\hspace{2.5mm}}c@{\hspace{2.5mm}}c@{\hspace{2.5mm}}c@{\hspace{2.5mm}}c@{\hspace{2.5mm}}c@{\hspace{2.5mm}}c@{\hspace{2.5mm}}c@{\hspace{2.5mm}}c@{\hspace{2.5mm}}c}
\toprule
\multicolumn{18}{c}{\textbf{Cartographer random forest ATE prediction RMSEs without and with feature permutation (all values in centimeters)}} \\ 
\midrule
\multirow{2}{1.2cm}{Dataset} & \multirow{2}{*}{Parameter} & \multicolumn{16}{c}{Data sequences} \\
 &  & 1 & 2 & 3 & 4 & 5 & 6 & 7 & 8 & 9 & 10 & 11 & 12 & 13 & 14 & 15 & 16 \\
\midrule
\multirow{8}{1.2cm}{PASTEL Velodyne VLP16 LiDAR} 
& RMSE w/o permutation       &   3.5 & 2.1 &   1.3 &   3.5 & 1.3 & 1.1 & 0.9 & 0.3 & 4.7 & 1.9 & 2.0 & 1.0 & 1.5 & 1.2 & 4.3 & 2.9 \\
\cmidrule{2-18}
&  \textbf{Rotation\_weight}     & \g4.9  & \g3.7 & \g1.9 &   7.3  & \g2.6 & \g1.7 & \g1.2 & \g0.4 &   9.3  & \g3.7 & \g2.8 & \g1.2 & \g2.0 & \g1.3 & \g4.9  &   5.0 \\
&  \textbf{Translation\_weight}  & \g4.5  & \g2.9 & \g3.0 & \g4.1  & \g2.0 & \g1.6 & \g2.4 & \g0.8 &   10.5 & \g3.3 &   4.2 & \g1.4 & \g2.2 & \g1.4 &   6.9  & \g4.1 \\
&  \textbf{High\_resolution}     & \g4.7  & \g3.6 & \g2.1 & \g4.1  & \g2.2 & \g2.1 & \g1.4 & \g0.8 &   8.4  & \g3.3 & \g3.0 & \g1.4 & \g1.9 & \g1.6 & \g4.6  & \g3.4 \\
&  \textbf{Low\_resolution}      &   11.3 &   5.4 &   3.6 & \g5.4  & \g2.5 & \g1.9 &   3.1 & \g1.0 &   15.3 &   5.7 &   6.5 & \g2.3 &   4.0 &   3.5 &   9.5  &   7.0 \\
&  \textbf{Submap\_size}         & \g3.6  & \g2.1 & \g1.3 & \g3.5  & \g1.4 & \g1.2 & \g1.0 & \g0.5 & \g5.6  & \g1.9 & \g2.1 & \g1.8 & \g1.8 & \g2.5 & \g5.4  & \g2.9 \\
&  \textbf{Voxel\_filter\_size}  & \g4.5  & \g3.8 & \g1.5 & \g4.2  & \g2.0 & \g1.5 & \g1.1 & \g0.4 & \g6.2  & \g2.7 & \g2.6 & \g1.3 & \g1.8 & \g1.3 & \g4.5  & \g4.2 \\
&  \textbf{LiDAR\_max\_range}    &   17.3 &   4.9 &   4.2 &   10.3 &   5.6 &   4.1 &   3.0 & \g0.7 &   13.3 &   6.8 &   5.5 & \g2.7 &   5.8 &   5.0 &   11.2 &   7.1 \\
\midrule
\multirow{8}{1.2cm}{PASTEL Ouster OS1-16 LiDAR} 
& RMSE w/o permutation       & 2.5 & 1.3 & 1.5 & 3.6 & 0.8 & 0.3 & 1.0 & 0.5 & 6.0 & 2.8 & 0.5 & 0.8 & 1.7 & 1.2 & 4.0 & 2.4 \\
\cmidrule{2-18}
& \textbf{Rotation\_weight}      & \g3.8  & \g2.6 & \g2.4 & \g5.3 & \g1.0 & \g0.5 & \g1.0 & \g0.6 & \g8.1  & \g4.4  & \g0.8 & \g0.9 & \g2.0 & \g1.4 & \g4.5 &   4.8 \\
& \textbf{Translation\_weight}   & \g4.1  & \g2.5 & \g2.3 & \g4.3 & \g0.8 & \g0.6 & \g1.4 & \g0.9 &   10.4 & \g3.8  & \g0.8 & \g0.9 & \g2.0 & \g1.3 &   6.2 &   4.7 \\
& \textbf{High\_resolution}      & \g3.6  & \g2.8 & \g2.3 & \g4.6 & \g1.1 & \g1.0 & \g1.1 & \g0.8 & \g8.1  & \g4.4  & \g0.8 & \g1.0 & \g1.9 & \g1.3 & \g4.9 & \g4.0 \\
& \textbf{Low\_resolution}       &   10.3 &   3.8 & \g2.5 &   6.6 & \g1.2 & \g0.8 & \g1.5 & \g1.1 &   11.9 &   9.0  & \g1.1 & \g1.6 & \g2.5 &   3.7 &   7.6 &   6.7 \\
& \textbf{Submap\_size}          & \g3.1  & \g2.0 & \g1.5 & \g3.8 & \g0.8 & \g0.4 & \g1.1 & \g0.7 & \g7.1  & \g3.3  & \g0.7 &   3.3 & \g2.0 &   3.5 & \g5.5 & \g2.5 \\
& \textbf{Voxel\_filter\_size}   & \g3.5  & \g2.1 & \g2.0 & \g4.4 & \g0.9 & \g0.5 & \g1.0 & \g0.6 &   9.5  & \g4.2  & \g0.8 & \g0.8 & \g1.9 & \g1.3 & \g4.1 & \g4.0 \\
& \textbf{LiDAR\_max\_range}     &   15.3 &   4.0 & \g3.4 &   9.3 & \g1.5 & \g0.7 & \g1.3 & \g0.8 &   11.8 &   10.9 & \g1.6 & \g2.8 & \g2.9 &   5.6 &   7.3 &   6.0 \\
\midrule
\multirow{8}{1.2cm}{PASTEL Ouster OS0-128 LiDAR} 
& RMSE w/o permutation         &   1.7 &   0.4 &   1.0 &   0.5 &   0.6 &   0.3 &   0.8 &   0.7 &   1.3 &   0.2 &   1.0 &   0.9 &   1.3 &    1.2 &   1.4 &   0.9 \\
\cmidrule{2-18}
& \textbf{Rotation\_weight}    & \g1.8 & \g0.4 & \g1.1 & \g0.8 & \g0.8 & \g0.6 & \g0.9 & \g0.8 & \g1.3 & \g0.3 & \g1.3 & \g1.0 & \g1.4 & \g1.4 & \g1.6 & \g1.0 \\
& \textbf{Translation\_weight} & \g1.8 & \g0.5 & \g1.3 & \g1.2 & \g0.7 & \g0.4 & \g1.5 & \g0.9 & \g1.4 & \g0.3 & \g1.5 & \g1.0 & \g1.4 & \g1.3 & \g1.7 & \g1.2 \\
& \textbf{High\_resolution}    & \g1.9 & \g0.7 & \g1.4 & \g1.2 & \g0.8 & \g0.9 & \g1.3 & \g1.0 & \g1.4 & \g0.5 & \g1.5 & \g1.0 & \g1.5 & \g1.8 & \g1.6 & \g1.6 \\
& \textbf{Low\_resolution }    & \g2.1 & \g0.7 & \g1.6 & \g1.2 & \g0.7 & \g1.3 & \g1.8 & \g1.2 & \g1.5 & \g0.5 & \g1.2 & \g1.3 & \g1.7 & \g2.6 & \g1.8 & \g2.4 \\
& \textbf{Submap\_size}        & \g2.2 & \g0.6 & \g1.2 & \g0.8 & \g0.7 & \g0.3 & \g1.0 & \g0.8 & \g1.4 & \g0.2 & \g1.2 &   2.9 & \g1.5 &   4.2 & \g2.6 & \g1.2 \\
& \textbf{Voxel\_filter\_size} & \g1.9 & \g0.5 & \g1.1 & \g0.6 & \g0.7 & \g0.3 & \g0.9 & \g0.9 & \g1.4 & \g0.2 & \g1.0 & \g0.9 & \g1.5 & \g1.4 & \g1.7 & \g1.1 \\
& \textbf{LiDAR\_max\_range}   & \g3.7 & \g1.0 & \g1.2 & \g1.2 & \g0.7 & \g0.7 & \g1.5 & \g1.1 & \g1.7 & \g0.3 & \g1.3 & \g2.1 & \g1.9 &   4.6 & \g2.2 &   5.1 \\
\midrule
\multirow{8}{1.2cm}{NTU VIRAL Ouster OS1-16 LiDAR}
& RMSE w/o permutation     & 4.5 & 13.0 & 2.1 & 2.4 & 4.5 & 6.9 & 1.1 & 3.0 & 6.4 & 12.3 & 17.9& 14.2 & \multicolumn{4}{c}{/} \\
\cmidrule{2-18}
& \textbf{Rotation\_weight}    & \g5.6  & \g17.9 & \g3.4  & \g3.7  & \g5.8  & \g9.1  & \g1.9 & \g4.6  & \g7.1  &   19.0 &   34.4 & \g19.0 & \multicolumn{4}{c}{/} \\
& \textbf{Translation\_weight} & \g5.3  & \g13.3 &   6.9  &   4.4  &   18.3 &   11.8 & \g1.3 &   5.6  & \g7.0  & \g14.4 & \g22.0 & \g15.1 & \multicolumn{4}{c}{/} \\
& \textbf{High\_resolution}    & \g5.8  & \g13.1 & \g3.4  & \g4.0  & \g5.2  & \g9.7  & \g1.8 & \g4.7  & \g7.4  & \g16.7 & \g21.4 & \g17.1 & \multicolumn{4}{c}{/} \\
& \textbf{Low\_resolution}     &   10.1 &   26.7 &   8.4  &   7.9  &   21.4 &   14.7 & \g2.8 &   10.5 &   17.8 &   35.3 &   81.0 &   26.9 & \multicolumn{4}{c}{/} \\
& \textbf{Submap\_size}        &   18.4 &   31.9 & \g2.6  & \g2.8  & \g5.4  & \g10.2 &   7.0 &   5.5  &   11.3 &   35.9 &   56.5 &   72.3 & \multicolumn{4}{c}{/} \\
& \textbf{Voxel\_filter\_size} & \g5.7  & \g13.6 & \g3.6  & \g3.8  & \g5.3  & \g7.3  & \g1.3 & \g3.8  & \g6.6  & \g13.6 & \g19.6 & \g14.3 & \multicolumn{4}{c}{/} \\
& \textbf{LiDAR\_max\_range}   &   26.8 &   33.6 &   21.3 &   12.5 &   21.1 &   29.9 &   5.7 &   13.7 &   34.7 &   32.2 &   36.2 & \g18.9 & \multicolumn{4}{c}{/} \\
\bottomrule
\end{tabular}
}
\end{table*}

\subsubsection{Permutation-based feature importance data analysis}
\label{subsubsec:carto_permutation_based_analysis}

We first differentiate between parameters with high and low influence using the data obtained from the random forest feature importance analysis. 
To determine whether a parameter's influence is low for a particular data sequence and LiDAR, we compare the RMSE value when permuting the value of each parameter to the models' base RMSE without permutation. We consider that a parameter has low influence for a particular data sequence and LiDAR if its permutation increases the model's base RMSE by less than 50\% or by less than 2 cm. The ATE values for low-influence cases have been printed in gray color in Table \ref{table:cartographer_permutation_importance} to help easily differentiate them from high influence cases.

This analysis can be used to find the most influential parameters. As shown, \textbf{LiDAR\_max\_range} has the strongest overall influence among all the analyzed combinations of sequence and LiDAR. It is influential in 
58\% (35/60) 
of cases, and in 
26 out of those 35 cases, 
it is the most influential parameter. The second most influential parameter is \textbf{Low\_resolution}, which is influential in 
50\% (30/60)
of the cases and the most influential parameter in 
8 cases. 
Most of the other parameters have a lower overall influence, but still have an influence in some cases. Parameter \textbf{Submap\_size} is even the most influential in NTU VIRAL sequences 7 and 12, and in PASTEL sequence 12 with Ouster LiDARs. IMU parameters \textbf{Rotation\_weight} and \textbf{Translation\_weight} are never the most influential parameters, but have a significant influence in 
10\% (6/60) and 18\% (11/60)
of the cases, respectively.

The parameters with the lowest overall influence are \textbf{Voxel\_filter\_size} and \textbf{High\_resolution}. They each have high influence in one case (PASTEL sequence 9 with Ouster OS1-16 and Velodyne VLP16 LiDARs respectively), 
but not as significant as the other more influential parameters.
Hence, we classify them as \textit{low-influence parameters} and exclude them from further analysis. As setting them to low values may increase processing time and memory expenditure, we recommend not tuning them and simply setting them to the higher considered values from Table \ref{table:grid_search}.

\subsubsection{Correlation data analysis - need for manual tuning}
\label{subsubsec:carto_correlation analysis_manual}
The parameters that were not classified as \emph{low-influence parameters} are further analyzed using correlation. Table \ref{table:cartographer_correlations} shows the correlations between each Cartographer parameter and the ATE values in the \textit{LIO performance data} for each sequence and LiDAR available. Cases where a parameter's correlation coefficient changes its sign in different sequences indicate that the parameter behaves differently depending on the situation, and that it requires manual tuning. This is the case for \textbf{LiDAR\_max\_range} and \textbf{Low\_resolution} with LiDARs Ouster OS1-16 and Velodyne VLP16. Manual tuning is also required in cases where the parameter correlations are close to zero, $(-0.2 < r < 0.2)$, (i.e., ambiguous influence on ATE), but have high influence on the permutation feature analysis. 
To facilitate the identification of such cases, the correlation values in Table \ref{table:cartographer_correlations} which correspond to low-influence cases identified in Table \ref{table:cartographer_permutation_importance} were printed in gray color, leaving the influential parameters printed in black. As shown in Table \ref{table:cartographer_correlations}, this is again the case for \textbf{LiDAR\_max\_range} and \textbf{Low\_resolution} in many sequences and LiDAR combinations, which further reinforces the need to tune them manually.

\def\sp{\hspace{2.25mm}}
\def\spI{\hspace{1mm}}
\begin{table*}[ht!]
\centering
\caption{Cartographer parameter correlations with ATE. Values corresponding to low sensitivity cases from Table \ref{table:cartographer_permutation_importance} are printed in gray. Cases with strong influence but weak correlations $(-0.2 < r < 0.2)$ indicate ambiguous parameter choices.
}
\label{table:cartographer_correlations}
\scalebox{0.9}{
\begin{tabular}{@{}l@{\spI}l@{\sp}|@{\sp}c@{\sp}c@{\sp}c@{\sp}c@{\sp}c@{\sp}c@{\sp}c@{\sp}c@{\sp}c@{\sp}c@{\sp}c@{\sp}c@{\sp}c@{\sp}c@{\sp}c@{\sp}c}
\toprule 
\multicolumn{18}{c}{\textbf{Cartographer parameter correlations with ATE}} \\
\midrule
\multirow{2}{*}{Dataset} & \multirow{2}{*}{Parameter} & \multicolumn{16}{c}{Data sequences} \\
 &  & 1 & 2 & 3 & 4 & 5 & 6 & 7 & 8 & 9 & 10 & 11 & 12 & 13 & 14 & 15 & 16 \\
\midrule
\multirow{7}{1.2cm}{PASTEL Velodyne VLP16 LiDAR} 
 & \textbf{Rotation\_weight}     &\g -0.12 & \g-0.25 & \g-0.17 &   -0.43 & \g-0.26 & \g-0.11 & \g-0.01 & \g-0.08 &  -0.11  & \g-0.26 & \g-0.11 & \g-0.06 & \g-0.07 & \g-0.02 & \g-0.04 & -0.22 \\
 & \textbf{Translation\_weight}  &\g 0.07  & \g0.06  & \g0.31  & \g0.09  & \g0.10  & \g0.13  & \g0.29  & \g0.26  &   0.14  & \g0.04  &   0.23  & \g0.13  & \g0.12  & \g0.04  &   0.19  & \g-0.03 \\
 & \textbf{High\_resolution}     &\g 0.01  & \g0.09  & \g0.21  & \g0.07  & \g0.10  & \g0.22  & \g0.12  & \g0.39  &  -0.01  & \g-0.01 & \g0.08  & \g0.17  & \g0.03  & \g0.08  & \g0.02  & \g-0.02 \\
 & \textbf{Low\_resolution}      &  -0.31  & -0.22   &   0.41  & \g-0.17 & \g-0.05 & \g0.17  &   0.19  & \g0.39  &  -0.28  &   -0.36 &   -0.38 & \g0.23  &   0.21  &   0.17  &   0.33  &  -0.31 \\
 & \textbf{Submap\_size}         & \g0.01  & \g-0.03 & \g-0.03 & \g-0.07 & \g0.01  & \g-0.06 & \g-0.07 & \g-0.11 & \g0.00  & \g0.05  & \g-0.02 & \g-0.21 & \g-0.04 & \g-0.18 & \g-0.11 & \g-0.01 \\
 & \textbf{Voxel\_filter\_size}  & \g-0.06 & \g-0.09 & \g-0.07 & \g-0.05 & \g-0.04 & \g-0.07 & \g0.01  & \g-0.04 & \g0.03  & \g0.04  & \g0.01  & \g-0.04 & \g-0.03 & \g-0.05 & \g-0.03  & \g0.00 \\
 & \textbf{LiDAR\_max\_range}    &   0.73  &   0.07  &   0.50  &   0.63  &   0.80  &   0.79  &  -0.17  & \g-0.04 &   0.23  &   0.55  &   0.03  & \g-0.39 &   -0.54 &   -0.47 &   -0.48 &   0.19 \\
 \midrule
\multirow{7}{1.2cm}{PASTEL Ouster OS1-16 LiDAR}
 & \textbf{Rotation\_weight}    & \g-0.10 & \g-0.10 & \g-0.24 & \g-0.29 & \g-0.17 & \g0.02  & \g0.09  & \g0.06  & \g-0.06 & \g-0.08 & \g-0.06 & \g-0.03 & \g0.01  & \g-0.05 & \g-0.02 & -0.24 \\
 & \textbf{Translation\_weight} & \g0.09  & \g0.14  & \g0.20  & \g0.02  & \g0.08  & \g0.02  & \g0.34  & \g0.25  &   0.19  & \g0.07  & \g0.04  & \g0.04  & \g0.04  & \g0.03  &   0.17  & -0.18 \\
 & \textbf{High\_resolution}    & \g-0.03 & \g0.20  & \g0.13  & \g0.00  & \g0.32  & \g0.61  & \g0.16  & \g0.25  & \g-0.09 & \g-0.07 & \g0.05  & \g0.08  & \g0.05  & \g0.02  & \g-0.04 & \g-0.05 \\
 & \textbf{Low\_resolution}     &  -0.29  &   0.00  & \g0.01  &   -0.32 & \g-0.05 & \g0.23  & \g0.43  & \g0.41  &   -0.18 &   -0.49 & \g0.03  & \g0.17  & \g0.25  &   0.06  &   0.26  & -0.37 \\
 & \textbf{Submap\_size}        & \g-0.05 & \g-0.10 & \g-0.03 & \g-0.01 & \g-0.04 & \g-0.03 & \g-0.18 & \g-0.16 & \g0.00  & \g0.01  & \g-0.07 &  -0.46  & \g-0.01 &  -0.29  & \g-0.12 & \g0.05 \\
 & \textbf{Voxel\_filter\_size} & \g-0.07 & \g-0.05 & \g-0.11 & \g-0.03 & \g-0.01 & \g-0.18 & \g-0.02 & \g-0.09 & -0.01   & \g0.01  & \g0.01  & \g-0.04 & \g-0.08 & \g-0.04 & \g0.00  & \g-0.06 \\
 & \textbf{LiDAR\_max\_range}   &   0.72  &  -0.28  & \g0.50  &   0.61  & \g0.54  & \g0.36  & \g-0.12 & \g0.00  &   0.22  &   0.67  & \g-0.53 & \g-0.24 & \g-0.37 &  -0.46  &  -0.27  & 0.19 \\
 \midrule
\multirow{7}{1.2cm}{PASTEL Ouster OS0-128 LiDAR} 
& \textbf{Rotation\_weight}    & \g-0.05 & \g0.00  & \g0.13  & \g0.21  & \g0.44  & \g0.31  & \g0.06  & \g0.05  & \g-0.03 & \g0.31  & \g0.03  & \g-0.01 & \g-0.10  & \g-0.06 & \g 0.01 & \g-0.01 \\
& \textbf{Translation\_weight} & \g-0.05 & \g-0.16 & \g0.18  & \g0.28  & \g-0.02 & \g-0.0  & \g0.30  & \g0.19  & \g-0.13 & \g-0.25 & \g-0.16 & \g-0.04 & \g-0.09  & \g-0.05 & \g-0.06 & \g0.01 \\
& \textbf{High\_resolution}    & \g0.15  & \g0.29  & \g0.35  & \g0.43  & \g0.32  & \g0.42  & \g0.28  & \g0.34  & \g0.01  & \g0.49  & \g-0.10 & \g0.05  & \g0.08   & \g0.08  & \g-0.02 & \g0.14 \\
& \textbf{Low\_resolution}     & \g0.13  & \g0.32  & \g0.34  & \g0.23  & \g0.22  & \g0.65  & \g0.34  & \g0.35  & \g0.17  & \g0.33  & \g0.20  & \g0.13  & \g 0.26  & \g0.22  & \g 0.16 & \g0.21 \\
& \textbf{Submap\_size}        & \g-0.17 & \g-0.19 & \g-0.08 & \g-0.18 & \g0.10  & \g0.05  & \g-0.15 & \g-0.15 & \g-0.01 & \g-0.12 & \g-0.10 &  -0.58  & \g-0.09  &   -0.44 & \g-0.47 & \g-0.11 \\
& \textbf{Voxel\_filter\_size} & \g-0.06 & \g-0.04 & \g0.00  & \g-0.10 & \g-0.08 & \g-0.04 & \g-0.05 & \g-0.04 & \g-0.03 & \g-0.01 & \g0.02  & \g-0.01 & \g-0.13  & \g-0.02 & \g 0.18 & \g-0.05 \\
& \textbf{LiDAR\_max\_range}   & \g-0.65 & \g-0.56 & \g-0.17 & \g-0.34 & \g-0.14 & \g-0.29 & \g-0.26 & \g-0.24 & \g-0.34 & \g0.03  & \g-0.37 & \g-0.33 & \g-0.37  &  -0.55  & \g-0.32 &  -0.75 \\
\midrule
\multirow{7}{1.2cm}{NTU VIRAL Ouster OS1-16 LiDAR} 
& \textbf{Rotation\_weight}     & \g0.02  & \g-0.13  & \g-0.03  & \g-0.01 & \g-0.06 & \g-0.05 & \g-0.07 & \g-0.11 & \g-0.03 &   -0.19 &   -0.21 & \g-0.11 & \multicolumn{4}{c}{/} \\
& \textbf{Translation\_weight}  & \g0.01  & \g0.05   &   0.20   &   0.18  &   0.33  &   0.16  & \g0.07  &   0.10  & \g0.06  & \g0.07  & \g-0.01 & \g0.01  & \multicolumn{4}{c}{/} \\
& \textbf{High\_resolution}     & \g0.01  & \g-0.01  & \g-0.01  & \g0.10  & \g0.01  & \g0.01  & \g0.07  & \g-0.02 & \g0.00  & \g-0.04 & \g-0.05 & \g0.02  & \multicolumn{4}{c}{/} \\
& \textbf{Low\_resolution}      &   -0.16 &   -0.39  &   -0.20  &   -0.38 &   0.27  &   -0.16 & \g-0.05 &   -0.30 &   -0.24 &   -0.55 &   -0.73 &   -0.18 & \multicolumn{4}{c}{/} \\
& \textbf{Submap\_size}         &   -0.23 &   -0.27  & \g0.01   & \g0.02  & \g0.02  & \g0.06  &  -0.40  &   -0.08 &   -0.09 &   -0.27 &   -0.36 &   -0.80 & \multicolumn{4}{c}{/} \\
& \textbf{Voxel\_filter\_size}  & \g0.00  & \g-0.01  & \g0.01   & \g0.00  & \g0.00  & \g-0.01 & \g0.02  & \g-0.02 & \g0.00  & \g0.01  & \g0.01  & \g0.01  & \multicolumn{4}{c}{/} \\
& \textbf{LiDAR\_max\_range}    &   0.67  &   0.29   &   0.84   &   0.70  &   -0.18 &   0.78  &   -0.14 &   0.52  &   0.81  &   -0.07 &   0.09  & \g-0.01  & \multicolumn{4}{c}{/} \\
 \bottomrule
\end{tabular}
}
\end{table*}

\subsubsection{Correlation data analysis - overarching trends}
\label{subsubsec:carto_correlation analysis_trends}

The other influential parameters (\textbf{Rotation\_weight},   \textbf{Translation\_weight}, and \textbf{Submap\_size}) also exhibit cases of weak correlations, but in a limited number of cases. That is the case for \textbf{Rotation\_weight} in PASTEL sequence 9 with Velodyne VLP16 ($r=-0.11$). In other cases in which \textbf{Rotation\_weight} has a high influence value in Table \ref{table:cartographer_correlations}, its correlations are larger and consistently negative (e.g., Velodyne VLP16 in PASTEL sequences 4 and 16). This leads us to recommend to set \textbf{Rotation\_weight} to a high value, keeping in mind that there are isolated cases where tuning it by hand may be beneficial. The situation is similar for \textbf{Submap\_size}, whose correlations are consistently negative for all cases in which it has a high influence (see Table \ref{table:cartographer_correlations}). The recommendation is to set it to the higher considered value from Table \ref{table:grid_search}. This parameter also has one isolated case where it has a strong influence and its correlation is weak (NTU VIRAL seq. 9).

Contrarily, \textbf{Translation\_weight} has more cases of significant influence and weak positive correlations (PASTEL seq. 9 with Velodyne VLP16, seq. 9 and 15 with Ouster OS1-16, and NTU VIRAL seq. 6 and 8). 
In the other cases where it has significant influence, its correlations are positive, indicating that lower parameter values are preferred. Hence, we recommend setting it to the lowest value from the grid search.

Although we recommended the \textbf{LiDAR\_max\_range} parameter for manual tuning in Section \ref{subsubsec:carto_correlation analysis_manual}, there are two overarching trends present in the correlation data which can facilitate the manual tuning process. 
Firstly, when processing the PASTEL dataset with the Ouster OS0-128 LiDAR, \textbf{LiDAR\_max\_range} has negative correlations in all sequences except in sequence 10. Furthermore, everywhere where the parameter's influence is strong, these correlation values are strong as well (e.g., $r < -0.5$). This indicates that Cartographer benefits from larger values of \textbf{LiDAR\_max\_range} when using the Ouster OS0-128 LiDAR, removing the need to manually tune it in this case. Secondly, in PASTEL sequences 3-6, which have very low distances from obstacles (see last three columns of Table \ref{table:combined_dataset}), the correlations corresponding to \textbf{LiDAR\_max\_range} are large and positive ($r>0.5$) when using the Velodyne VLP16 and Ouster OS1-16 LiDARs. This indicates that Cartographer benefits from reducing \textbf{LiDAR\_max\_range} in such cases. Per the LiDAR's technical characteristics (available in \cite{Milijas2025}), these devices have a low vertical field-of-view ($30\degree$), and 16 LiDAR beams, while the Ouster OS0-128 has 128 beams and a large vertical field-of-view ($90\degree$). We believe that this difference in LiDAR characteristics causes this difference in performance and recommend reducing \textbf{LiDAR\_max\_range} when flying in confined spaces with 16-beam or low vertical-field-of-view LiDARs.

\subsubsection{Cartographer tuning recommendations}
\label{subsubsec:carto_recommendations}

\begin{table}[pos=ht!]
    \caption{Summary of Cartographer tuning recommendations and recommended starting tuning values. The parameters are arranged by the need for manual tuning in decreasing order. Parameter values such as \textit{low} or \textit{high} correspond to values in Table \ref{table:grid_search}.
    }
    \label{table:cartographer_recommendations}
    \centering
    \scalebox{0.9}{
    \begin{tabular}{@{\hspace{1mm}}l@{\hspace{2mm}}l}
    \toprule
    \multicolumn{2}{c}{\textbf{Cartographer tuning recommendations}} \\
    \midrule
    \multirow{6}{*}{\textbf{LiDAR\_max\_range}}
        & \multirow{6}{5.75cm}{Use lower values (e.g. \bl{20} m) when using low field-of-view LiDARs in confined spaces, and high values (e.g. 50 m) with large field-of-view LiDARS. Manually tune otherwise, starting from medium value (e.g. 35 m).} \\ \\ \\ \\ \\ \\
        \midrule
    \multirow{4}{*}{\textbf{Low\_resolution}}   
        & \multirow{4}{5.75cm}{Start from a higher value (e.g. 50 cm) and reduce if necessary. Keep computational constraints in mind when executing on low-resource hardware.} \\ \\ \\ \\
        \midrule
    \multirow{2}{*}{\textbf{Translation\_weight}}   
        & \multirow{2}{5.75cm}{Low values (e.g. 5) expected to perform well.} 
        \\ \\
        \midrule
    \multirow{2}{*}{\textbf{Rotation\_weight}}   
        & \multirow{2}{5.75cm}{High values (e.g. 100) expected to perform well. 
        } 
        \\ \\ 
        \midrule
    \multirow{2}{*}{\textbf{Submap\_size}}  
        & \multirow{2}{5.75cm}{High values (e.g. 1000) expected to perform well.} \\ \\
        \midrule
    \multirow{3.5}{*}{\textbf{High\_resolution,}} 
        & \multirow{5}{5.75cm}{Parameters have low influence in the data sequences analyzed in this work. Set to higher value (e.g. 20 cm / 15 cm) as they influence processing time and memory consumption.} \\ \\
    \multirow{3}{*}{\textbf{Voxel\_filter\_size}}
        \\ \\ \\
    \bottomrule 
    \end{tabular}
    } 
\end{table}

The above analyses are distilled into the parameter tuning recommendations shown in Table \ref{table:cartographer_recommendations}. These recommendations are guidelines for the parameters requiring manual tuning and suggest values for the other parameters in order to achieve good performance in scenarios similar to the PASTEL and NTU VIRAL datasets. The parameters are arranged by the expected need for manual tuning in decreasing order. \textbf{LiDAR\_max\_range} and \textbf{Low\_resolution}, which have shown high permutation-based feature importance and need for manual tuning have the higher rank in the recommendations. They are followed by the rest of the influential parameters  (\textbf{Translation\_weight}, \textbf{Rotation\_weight}, and \textbf{Submap\_size}), whose general trends were examined in Section \ref{subsubsec:carto_correlation analysis_trends}. These trends are reflected in the recommendations in Table \ref{table:cartographer_recommendations}. The \textit{low-influence parameters} (\textbf{High\_resolution} and \textbf{Voxel\_filter\_size}) share the last rank in the recommendations. The table also contains recommended starting values for each of the parameters.

When deploying Cartographer in scenarios similar to those of the analyzed datasets, we recommend the practitioner to first set the recommended values from the table and assess the system performance. If the performance is not satisfactory, the parameters should be fine-tuned as recommended in Table \ref{table:cartographer_recommendations} starting from the top of the table. We expect that a simplified tuning process focusing only on the two most influential parameters and leaving the others at the starting values will often provide satisfactory results.

\subsubsection{Validation of Cartographer tuning recommendations}
\label{subsubsec:carto_validation}

This section validates the proposed parameter tuning recommendations using the \textit{LIO performance data}. We confirm that using the proposed starting values for the parameters in Table \ref{table:cartographer_recommendations} and tuning only the two most influential parameters designated for manual tuning (\textbf{LiDAR\_max\_range} and \textbf{Low\_resolution}), Cartographer obtains outcomes similar to the optimal ATE values found by the grid search (Table \ref{table:best_grid_search_ATEs}). It also validates that the tuning order recommended by Table \ref{table:cartographer_recommendations} is indeed the order of falling importance for manual tuning. 

\begin{figure}[pos=ht!]
    \centering
    \includegraphics[width=0.99\linewidth, trim={5.1cm, 8.0cm, 5.80cm, 9.150cm}, clip]{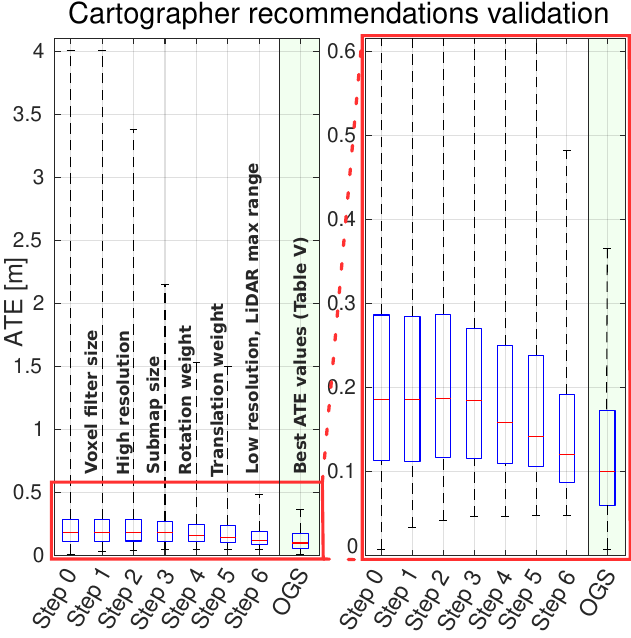} 
    \caption{
    Cartographer ATE distributions considering all sequence-LiDAR combinations in \emph{LIO performance data} along different validation steps. Step 0 considers all combinations of parameter values in Table \ref{table:grid_search}. From Step 1 to Step 5, a parameter (in reverse order to the tuning importance expressed in Table \ref{table:cartographer_recommendations}) is set to its recommended value. Steps 1 and 2, which reflect the influence of \textbf{Voxel\_filter\_size} and \textbf{High\_resolution}, have minimal influence on the ATE distributions, while the changes in the ATE distributions increase in Steps 3-5, hence validating the ordering of the parameters in the tuning recommendations. Step 6 manually tunes \textbf{LiDAR\_max\_range} and \textbf{Low\_resolution} for each of the LiDAR and data sequence combinations. For comparison, the ATE distribution from the optimal grid search (OGS) tuning (shown in Table \ref{table:grid_search}) is depicted in green. The ATE distribution resulting of the proposed tuning recommendations (Step 6) is similar (the quartiles have within 5 cm of difference) to the ATE distribution resulting from the optimal grid search, validating that satisfactory performance can be achieved following the proposed recommendations and manually tuning solely two parameters: \textbf{LiDAR\_max\_range} and \textbf{Low\_resolution}.
    }
    \label{fig:Cartographer_recommendation_validation}
\end{figure}

We test both through a validation process, the results of which are depicted in Figure \ref{fig:Cartographer_recommendation_validation}. The process consists of applying in 6 steps the parameter recommended starting values shown in Table \ref{table:cartographer_recommendations} one by one in reverse order (i.e. with rising importances) and observing the changes of the combined ATE distributions for all 60 LiDAR and data sequence combinations from the \textit{LIO performance data}. Step 0 in the process starts by considering all combinations of parameter values in Table \ref{table:grid_search} for all the LiDAR-sequence combinations. The resulting ATE distribution is shown in Figure \ref{fig:Cartographer_recommendation_validation}-Step 0. At each consecutive step, a parameter (in reverse order to the tuning importance expressed in Table \ref{table:cartographer_recommendations}) is set to its recommended value, and Figure \ref{fig:Cartographer_recommendation_validation} shows the resulting ATE distribution. For instance, in Step 1 \textbf{Voxel\_filter\_size} is fixed to 15 cm (recommended value from Table \ref{table:cartographer_recommendations}), and Figure \ref{fig:Cartographer_recommendation_validation}-Step 1 shows the ATE distribution considering the combinations of values for all parameters except \textbf{Voxel\_filter\_size}. 
It can be seen in Figure \ref{fig:Cartographer_recommendation_validation} that in Step 1 the minimum of the resulting ATE distribution moved upwards, without the quartiles or median changing. In Step 2, the next parameter in reverse order in Table \ref{table:cartographer_recommendations} (\textbf{High\_resolution}) is set to its recommended value. The process continues to Step 6, in which \textbf{Low\_resolution} and \textbf{LiDAR\_max\_range} are manually tuned. Analyzing the resulting ATE distributions in Figure \ref{fig:Cartographer_recommendation_validation}, it can be observed that the \emph{low-influence parameters} --\textbf{Voxel\_filter\_size} (Step 1) and \textbf{High\_resolution} (Step 2)-- have a weak influence on Cartographer performance, as their selection did not significantly alter the ATE distributions. The changes in the ATE distribution became stronger when the validation procedure reaches parameter \textbf{Submap\_size} (Step 3).

In Step 6 the two most influential parameters (\textbf{LiDAR\_ max\_range} and \textbf{Low\_resolution}) are manually tuned. In a scenario where ground truth infrastructure is unavailable, a human operator would perform manual tuning through visual inspection of the SLAM process and by analyzing the resulting map. In this work, since ground truth is available, manual tuning was performed by selecting the combination of values from Table \ref{table:grid_search} for both parameters that obtained the lowest ATE for each data sequence and LiDAR. The resulting ATE distribution after manual tuning is shown in Figure \ref{fig:Cartographer_recommendation_validation}-Step 6. For comparison, Figure \ref{fig:Cartographer_recommendation_validation}-OGS (marked in green) shows the ATE distribution resulting from the optimal tuning of parameters performed using grid search, which ATE values for each data sequence and LiDAR are shown in Table \ref{table:best_grid_search_ATEs}.
As shown, the ATE distribution after following the proposed recommendations (Step 6) is very close to the ATE distribution of the optimal tuning found through grid search. The minimum, median and the quartiles of the ATE distribution with optimal tuning are less than 5 cm lower than with the proposed tuning recommendations, while the maxima of the distributions show a difference of approximately 12 cm. 

To allow a deeper comparison, the resulting ATE values for each sequence and LiDAR obtained with the proposed tuning recommendations (after Step 6) are shown in Table \ref{table:cartographer_recommendation_validation} and compared to the ATE values obtained from grid search optimal tuning that were shown in Table \ref{table:best_grid_search_ATEs} (repeated in Table \ref{table:cartographer_recommendation_validation} for clarity). Notably, the proposed very simple tuning procedure obtains results within 5 cm from the grid-search-optimal values in 55/60 cases. All data sequence and LiDAR combinations obtain ATE values lower than 26 cm except NTU VIRAL sequence 11 which stands out with 48 cm. That same sequence also has the largest minimum ATE in the Cartographer grid search (36.5 cm in Table \ref{table:best_grid_search_ATEs}) indicating that it poses a particularly challenging scenario for Cartographer.

Based on the presented results, we conclude that the Cartographer tuning recommendations given in Table \ref{table:cartographer_recommendations} are valid for the considered scenarios.

\begin{table}[pos=ht!]
\centering
\caption{
Cartographer ATE values for each sequence and LiDAR obtained using: i) the parameters resulting from the proposed tuning recommendations (after Step 6), and ii) the grid search optimal parameters (OGS) shown in Table \ref{table:best_grid_search_ATEs} but repeated here for ease of comparison. All values are expressed in centimeters. 
}
\label{table:cartographer_recommendation_validation}
\scalebox{0.9}{
\begin{tabular}{@{\hspace{1mm}}l@{\hspace{2mm}}|c@{\hspace{2mm}}c|c@{\hspace{2mm}}c|c@{\hspace{2mm}}c}
\toprule
\multicolumn{7}{c}{\textbf{Individual ATE values from Cartographer validation Step 6}} \\
\multicolumn{7}{c}{\textbf{compared to optimal ATE values (OGS) from Table \ref{table:best_grid_search_ATEs}.}} \\ 
\multicolumn{7}{c}{\textbf{All values in centimeters.}} \\ 
\midrule
\mrIII{\textbf{Data sequence}} & \multicolumn{2}{c|}{\textbf{Velodyne}}  & \multicolumn{2}{c|}{\textbf{Ouster}} & \multicolumn{2}{c}{\textbf{Ouster}} \\
                      & \multicolumn{2}{c|}{\textbf{VLP16}}     & \multicolumn{2}{c|}{\textbf{OS1-16}} & \multicolumn{2}{c}{\textbf{OS0-128}} \\
             & \textbf{Step 6} & \textbf{OGS} & \textbf{Step 6} & \textbf{OGS} & \textbf{Step 6} & \textbf{OGS} \\   
\midrule              
PASTEL    1  & 12.6 &  10.0 & 17.9 & 13.8   &  11.5 &  9.0  \\
PASTEL    2  & 4.8  &  3.6  & 6.8  &  5.6   &  6.8  &  5.3  \\
PASTEL    3  & 4.9  &  4.5  & 9.2  &  7.7   &  7.4  &  5.7  \\
PASTEL    4  & 4.7  &  2.4  & 7.8  &  7.1   &  8.3  &  6.3  \\
PASTEL    5  & 5.4  &  4.9  & 10.2 &  2.5   &  9.9  &  6.5  \\
PASTEL    6  & 5.1  &  4.4  & 11.5 &  9.9   &  10.9 &  7.7  \\
PASTEL    7  & 7.2  &  6.6  & 13.5 & 12.8   &  13.7 & 11.8  \\
PASTEL    8  & 6.7  &  5.7  & 11.3 & 10.0   &  9.5  &  8.0  \\
PASTEL    9  & 6.0  &  5.0  & 9.7  &  8.8   &  8.5  &  5.6  \\
PASTEL    10 & 6.2  &  5.7  & 11.3 & 10.1   &  10.6 &  9.5  \\
PASTEL    11 & 9.4  &  0.8  & 17.9 & 16.2   &  18.3 &  2.8  \\
PASTEL    12 & 10.6 &  8.7  & 16.8 & 16.5   &  15.4 & 13.1  \\
PASTEL    13 & 13.7 &  10.5 & 22.8 & 12.5   &  22.9 & 18.4  \\
PASTEL    14 & 8.9  &  8.0  & 14.0 & 13.5   &  13.2 & 11.8  \\
PASTEL    15 & 12.4 & 11.8  & 24.7 & 24.0   &  23.6 & 18.7  \\
PASTEL    16 & 9.8  &  8.2  & 18.2 & 15.5   &  18.0 & 17.0  \\
\midrule             
NTU VIRAL 1  & \multicolumn{2}{c|}{/} & 22.9 & 21.1 & \multicolumn{2}{c}{/} \\
NTU VIRAL 2  & \multicolumn{2}{c|}{/} & 23.9 & 22.1 & \multicolumn{2}{c}{/} \\
NTU VIRAL 3  & \multicolumn{2}{c|}{/} & 18.1 & 17.7 & \multicolumn{2}{c}{/} \\
NTU VIRAL 4  & \multicolumn{2}{c|}{/} & 23.6 & 21.4 & \multicolumn{2}{c}{/} \\
NTU VIRAL 5  & \multicolumn{2}{c|}{/} & 20.1 & 18.6 & \multicolumn{2}{c}{/} \\
NTU VIRAL 6  & \multicolumn{2}{c|}{/} & 24.3 & 22.0 & \multicolumn{2}{c}{/} \\
NTU VIRAL 7  & \multicolumn{2}{c|}{/} & 23.3 & 21.5 & \multicolumn{2}{c}{/} \\
NTU VIRAL 8  & \multicolumn{2}{c|}{/} & 23.4 & 22.0 & \multicolumn{2}{c}{/} \\
NTU VIRAL 9  & \multicolumn{2}{c|}{/} & 21.6 & 20.3 & \multicolumn{2}{c}{/} \\
NTU VIRAL 10 & \multicolumn{2}{c|}{/} & 25.2 & 21.9 & \multicolumn{2}{c}{/} \\
NTU VIRAL 11 & \multicolumn{2}{c|}{/} & 48.1 & 36.5 & \multicolumn{2}{c}{/} \\
NTU VIRAL 12 & \multicolumn{2}{c|}{/} & 25.1 & 23.4 & \multicolumn{2}{c}{/} \\
\bottomrule
\end{tabular}
} 
\end{table}

\subsection{FAST-LIO2 parameter analysis}
\label{subsec:fastlio2_analysis}

Analogously to Cartographer, the influence of different \FL2 parameters is analyzed through the permutation-based random forest feature importances from Table \ref{table:fastlio2_permutation_importance} and correlation data from Table \ref{table:fastlio2_correlations}. First, we identify \textit{low-influence parameters} using the permutation-based feature importance analysis, and then identify the need for manual tuning and find overarching trends using the correlation data.

\begin{table*}[ht!]
\centering
\caption{
Random forest feature permutations analysis for \FL2 parameters for each sequence and LiDAR in the considered datasets. 
Cases which increase the RMSE by less than 50\% of the RMSE without permutation or by less than 2cm are marked in gray. All values are reported in centimeters.
}
\label{table:fastlio2_permutation_importance}
\scalebox{0.890}{
\begin{tabular}{@{\hspace{1mm}}l@{\hspace{1mm}}l@{\hspace{1mm}}|@{\hspace{1mm}}c@{\hspace{2.25mm}}c@{\hspace{2.25mm}}c@{\hspace{2.25mm}}c@{\hspace{2.25mm}}c@{\hspace{2.25mm}}c@{\hspace{2.25mm}}c@{\hspace{2.25mm}}c@{\hspace{2.25mm}}c@{\hspace{2.25mm}}c@{\hspace{2.25mm}}c@{\hspace{2.25mm}}c@{\hspace{2.25mm}}c@{\hspace{2.25mm}}c@{\hspace{2.25mm}}c@{\hspace{2.25mm}}c@{}}
\toprule
\multicolumn{18}{c}{\textbf{FAST-LIO2 random forest ATE prediction RMSEs without and with feature permutation. All values in centimeters.}} \\
\midrule
\multirow{2}{*}{Dataset} & \multirow{2}{*}{Parameter} & \multicolumn{16}{c}{Data sequences} \\
 &  & 1 & 2 & 3 & 4 & 5 & 6 & 7 & 8 & 9 & 10 & 11 & 12 & 13 & 14 & 15 & 16 \\
\midrule
\multirow{7}{1.2cm}{PASTEL Velodyne VLP16} 
& RMSE w/o permutation          &    4.1 &   30.0 &   22.4 &   63.9 &   50.4 &   16.0 &    10.1 &   33.4 &   8.1 &   32.8 &     28.4 &    36.7 &    33.2 &    47.4 &    37.6 &    40.9 \\
\cmidrule{2-18}                  
& \textbf{Acceleration\_covariance} &   23.6 &   106.5 &   76.0 &   120.8 &   87.9 &   67.2 &   23.6 &   131.4 &   37.8 & \g43.3 &   99.4  &   101.3 &   80.7 &   161.4 &   134.3 &   107.7 \\
& \textbf{Gyroscope\_covariance}    & \g4.2  & \g30.6  & \g24.4 & \g63.9  & \g53.0 & \g17.6 & \g11.7 & \g34.5  & \g8.1  & \g33.0 & \g30.2  & \g37.0  & \g33.7 & \g48.3  & \g41.9  & \g42.9  \\
& \textbf{Point\_filter\_number}    &   17.1 &   95.4  &   34.8 &   96.7  & \g68.6 &   43.8 &   20.5 &   116.3 &   20.8 & \g34.5 &   77.8  &   84.4  &   62.9 &   115.2 &   110.6 &   97.9  \\
& \textbf{Filter\_size\_surfaces}   &   7.8  &   47.7  &   64.2 &   152.6 &   96.6 &   70.4 &   30.9 &   139.1 &   27.7 & \g39.4 &   74.5  &   74.4  & \g46.5 &   126.2 &   100.8 &   83.9  \\
& \textbf{Filter\_size\_map}        &   19.5 &   98.0  &   85.0 &   144.4 &   90.8 &   76.6 &   32.1 &   209.4 &   32.1 & \g42.9 &   110.7 &   109.0 &   83.5 &   156.2 &   125.7 &   121.7 \\
& \textbf{LiDAR\_max\_range}        &   34.3 &   171.6 &   59.1 &   97.0  & \g74.1 &   61.0 &   21.2 &   75.0  &   37.7 & \g40.8 &   99.6  &   70.5  &   51.9 &   111.5 &   124.1 &   137.5 \\
\midrule
\multirow{7}{1.2cm}{PASTEL Ouster OS1-16 LiDAR}
& RMSE w/o permutation          &   7.0 &   37.5 &   58.6  &   78.3 &   13.8 &   24.9 &    7.6 &    37.6 &   11.6 &     4.1 &  17.9 &   29.2 &   26.0 &   51.7 &   18.7 &   49.5 \\
\cmidrule{2-18}  
& \textbf{Acceleration\_covariance} &   26.5 &   112.8 &   94.9  & \g114.4 &   45.2 &   77.5 &   13.3 &   150.1 &   19.5 &   14.7 &   65.4 &   111.1 &   58.4 &   149.5 &   131.5 &   101.7 \\
& \textbf{Gyroscope\_covariance}    & \g7.1  & \g39.7  & \g59.6  & \g78.5  & \g14.4 & \g25.6 & \g8.6  & \g41.3  & \g12.3 & \g4.8  & \g18.0 & \g29.6  & \g27.8 & \g53.3  & \g19.7  & \g50.2  \\
& \textbf{Point\_filter\_number}    &   21.1 &   109.8 & \g72.3  & \g85.2  &   26.9 &   51.8 & \g8.6  &   95.1  & \g16.1 &   11.1 &   48.5 &   105.0 & \g35.0 &   121.9 &   107.6 &   88.1  \\
& \textbf{Filter\_size\_surfaces}   & \g8.1  & \g51.3  &   89.5  &   131.3 &   36.4 &   57.6 &   17.6 &   147.6 & \g13.2 &   6.5  &   39.5 &   52.3  & \g36.0 &   86.8  &   68.6  &   94.8  \\
& \textbf{Filter\_size\_map}        &   22.0 &   115.2 &   107.2 &   120.0 &   50.3 &   81.7 &   19.5 &   170.3 &   18.8 &   13.7 &   70.7 &   94.0  &   53.4 &   162.0 &   124.5 &   135.1 \\
& \textbf{LiDAR\_max\_range}        &   30.7 &   173.0 & \g69.1  & \g89.9  &   29.3 & \g31.0 &   12.8 & \g45.4  &   21.8 &   13.8 &   73.3 &   64.1  &   47.1 &   98.4  &   119.8 &   113.4 \\
\midrule
\multirow{7}{1.2cm}{PASTEL Ouster OS0-128 LiDAR} 
& RMSE w/o permutation          &   10.1 &   39.9 &   27.6 &   44.9  & 10.6   & 6.1  &  3.3 &  31.4   & 22.2   &  0.3  & 23.4   & 18.1   & 13.4   & 33.4    & 42.8    & 37.0 \\
\cmidrule{2-18}
& \textbf{Acceleration\_covariance} &   31.5 &   92.9 &   65.4 &   104.3 &   17.0 &   20.5 &   6.8 &   136.4 & \g22.9 & \g0.8 &   42.8 &   61.7 &   27.8 &   103.8 &   126.1 &   89.5 \\
& \textbf{Gyroscope\_covariance}    & \g10.2 & \g40.4 & \g27.9 & \g48.4  & \g10.8 & \g6.4  & \g3.4 & \g33.7  & \g22.2 & \g0.3 & \g23.6 & \g18.4 & \g13.5 & \g33.7  & \g43.0  & \g39.1 \\
& \textbf{Point\_filter\_number}    &   23.1 &   78.3 & \g36.4 & \g48.2  & \g11.2 &   13.0 & \g4.8 &   94.0  & \g22.7 & \g0.5 &   39.3 &   35.6 & \g19.8 &   71.9  &   71.6  &   58.2 \\
& \textbf{Filter\_size\_surfaces}   &   19.9 &   60.5 &   48.5 &   71.6  & \g14.6 &   14.4 &   8.0 &   158.4 & \g22.9 & \g0.9 & \g34.9 &   46.0 & \g18.2 &   91.6  &   89.0  &   58.1 \\
& \textbf{Filter\_size\_map}        &   18.5 & \g57.0 &   57.4 &   96.9  & \g13.6 &   18.0 &   7.7 &   148.1 & \g22.2 & \g1.4 &   42.9 &   52.2 &   22.9 &   133.1 &   123.7 &   80.8 \\
& \textbf{LiDAR\_max\_range}        &   33.6 &   68.4 & \g41.1 & \g47.8  & \g12.9 &   10.2 &   5.6 & \g35.0  & \g22.6 & \g0.6 &   41.1 & \g22.9 & \g20.1 &   54.1  &   72.4  &   61.9 \\
\midrule
\multirow{7}{1.2cm}{NTU VIRAL Ouster OS1-16 LiDAR} 
& RMSE w/o permutation          &   11.6 &    29.0 &   0.2 &   0.1 &   0.2 &   0.2 &   3.2 &    9.7 &   17.0 &    9.8 &   15.5 &   9.6 & \multicolumn{4}{c}{/} \\
\cmidrule{2-18} 
& \textbf{Acceleration\_covariance} &   17.9 & \g31.4 & \g0.2 & \g0.3 & \g0.5 & \g0.3 & \g4.0 & \g9.7  & \g24.2  &   18.0 & \g16.3 & \g13.1 & \multicolumn{4}{c}{/} \\
& \textbf{Gyroscope\_covariance}    & \g13.2 & \g34.4 & \g0.2 & \g0.1 & \g0.2 & \g0.2 & \g3.5 & \g9.7  & \g17.1  & \g10.8 & \g18.5 & \g10.2 & \multicolumn{4}{c}{/} \\
& \textbf{Point\_filter\_number}    &   38.7 & \g36.4 & \g0.5 & \g0.9 & \g1.2 & \g1.2 &   7.1 & \g11.4 &   133.5 &   23.6 & \g21.9 & \g10.2 & \multicolumn{4}{c}{/} \\
& \textbf{Filter\_size\_surfaces}   &   24.8 & \g30.2 & \g0.4 & \g1.1 & \g1.1 & \g0.9 & \g4.7 & \g10.0 &   28.5  &   32.3 & \g19.4 & \g13.9 & \multicolumn{4}{c}{/} \\
& \textbf{Filter\_size\_map}        &   43.1 & \g37.7 & \g0.5 & \g1.4 & \g1.5 & \g1.1 &   6.9 & \g10.6 &   86.2  &   42.4 &   32.3 &   17.1 & \multicolumn{4}{c}{/} \\
& \textbf{LiDAR\_max\_range}        &   43.7 & \g38.2 & \g0.9 & \g0.7 & \g1.2 & \g0.8 &   6.8 & \g10.4 &   70.9  &   39.9 &   26.0 & \g13.3 & \multicolumn{4}{c}{/} \\
\bottomrule
\end{tabular}
}
\end{table*}

\subsubsection{Permutation-based feature importance data analysis}
\label{subsubsec:fl2_feature_importance}
Observing Table \ref{table:fastlio2_permutation_importance}, it can be noticed that all parameters except \textbf{Gyroscope\_covariance} have a significant influence in more than half of the analyzed cases. \textbf{Gyroscope\_covariance} had a negligible influence in all the used dataset-LiDAR combinations. In a majority of the cases the permutation of the parameter increased the random forest RMSEs in the order of millimeters, and the correlation coefficients were close to 0.0 on both datasets. We can conclude that this is a \emph{low-influence parameter} and that it can be left on its default value in scenarios with characteristics similar to those in the NTU VIRAL and PASTEL datasets.

\subsubsection{Correlation data analysis - need for manual tuning}
\label{subsubsec:fl2_correlation_manual}

\def\sp{\hspace{2.20mm}}
\begin{table*}[ht!]
\centering
\caption{
\FL2 parameter correlations with ATE. Values corresponding to low sensitivity cases from Table \ref{table:fastlio2_permutation_importance} are printed in gray. Cases with strong influence but weak correlations $(-0.2 < r < 0.2)$ indicate ambiguous parameter choices.}
\label{table:fastlio2_correlations}
\scalebox{0.890}{
\begin{tabular}{@{}l@{\hspace{1mm}}l@{\hspace{1mm}}|c@{\sp}c@{\sp}c@{\sp}c@{\sp}c@{\sp}c@{\sp}c@{\sp}c@{\sp}c@{\sp}c@{\sp}c@{\sp}c@{\sp}c@{\sp}c@{\sp}c@{\sp}c@{}}
\toprule
\multicolumn{18}{c}{\textbf{FAST-LIO2 parameter correlations with ATE}} \\
\midrule
\multirow{2}{*}{Dataset} & \multirow{2}{*}{Parameter} & \multicolumn{16}{c}{Data sequences} \\
 &  & 1 & 2 & 3 & 4 & 5 & 6 & 7 & 8 & 9 & 10 & 11 & 12 & 13 & 14 & 15 & 16 \\
\midrule
\multirow{6}{1.2cm}{PASTEL Velodyne VLP16 LiDAR} 
 & \textbf{Acceleration\_covariance} &   -0.29  &   -0.28  &   -0.12  &   -0.13  &   -0.24  &   -0.30  &   -0.17  &   -0.19  &   -0.22  & \g-0.12  &  -0.32   &   -0.43  &   -0.34  &   -0.31  &   -0.32  &   -0.31 \\
 & \textbf{Gyroscope\_covariance}    & \g0.00   & \g0.01   & \g0.03   & \g-0.01  & \g-0.03   & \g0.00   & \g0.00   & \g0.00   & \g0.01   & \g0.04   & \g-0.01  & \g0.00   & \g-0.01  & \g0.01   & \g0.00   & \g0.00  \\
 & \textbf{Point\_filter\_number}    &   0.20   &   0.23   &   -0.08  &   -0.07  & \g0.08   &   0.11   &   0.15   &   0.09   &   0.13   & \g0.04   &   0.12   &   0.36   &   0.27   &   0.21   &   0.18   &   0.29  \\
 & \textbf{Filter\_size\_surfaces}   &   0.05   &   0.00   &   0.20   &   0.47   &   0.28   &   0.35   &   0.24   &   0.29   &   0.17   & \g0.12   &   0.13   &   0.21   & \g0.16   &   0.29   &   0.16   &   0.25  \\
 & \textbf{Filter\_size\_map}        &   0.19   &   0.25   &   0.29   &   0.38   &   0.20   &   0.32   &   0.32   &   0.57   &   0.17   & \g0.16   &   0.34   &   0.30   &   0.35   &   0.41   &   0.33   &   0.40  \\
 & \textbf{LiDAR\_max\_range}        &   -0.50  &   -0.66  &   -0.13  &   -0.07  & \g-0.13  &   -0.19  &   -0.13  &   -0.03   &   -0.24  & \g-0.11 &  -0.29   &   -0.22  &   -0.26  &   -0.19  &   -0.29  &   -0.46 \\
\midrule
\multirow{6}{1.2cm}{PASTEL Ouster OS1-16 LiDAR} 
 & \textbf{Acceleration\_covariance} &   -0.27  &   -0.28   &   -0.22  & \g-0.29  &   -0.30  &   -0.35  &   -0.22  &   -0.28  &   -0.25  &   -0.23  &   -0.31  &   -0.53  &   -0.33  &   -0.38  &   -0.36  &   -0.36 \\
 & \textbf{Gyroscope\_covariance}    & \g-0.01   & \g-0.01  & \g-0.05  & \g-0.01  & \g-0.01  & \g-0.01   & \g0.00   & \g0.01   & \g0.00   & \g0.00   & \g0.02   & \g0.01   & \g0.03  & \g0.00    & \g0.00   & \g0.00  \\
 & \textbf{Point\_filter\_number}    &   0.25   &   0.28    & \g-0.04  & \g0.05   &   0.16   &   0.19   & \g0.10   &   0.01   & \g0.17   &   0.16   &   0.15   &   0.48   & \g0.21  &   0.25    &   0.26   &   0.33  \\
 & \textbf{Filter\_size\_surfaces}   & \g0.01   & \g0.00    &   0.19   &   0.47   &   0.23   &   0.28   &   0.25   &   0.31   & \g0.15   &   0.08   &   0.07   &   0.07   & \g0.16   &   0.15   &   0.12   &   0.23  \\
 & \textbf{Filter\_size\_map}        &   0.21   &   0.25    &   0.35   &   0.27   &   0.40   &   0.46   &   0.39   &   0.51   &   0.21   &   0.20   &   0.39   &   0.31   &   0.29   &   0.42   &   0.30   &   0.51  \\
 & \textbf{LiDAR\_max\_range}        &   -0.42  &   -0.65   & \g-0.03  & \g-0.03  &   -0.13  & \g-0.10  &   -0.14  & \g-0.08  &   -0.27  &   -0.23  &   -0.31  &   -0.16  &   -0.27  &   -0.18  &   -0.31  &   -0.40 \\
\midrule
\multirow{6}{1.2cm}{PASTEL Ouster OS0-128 LiDAR} 
 & \textbf{Acceleration\_covariance}  &    -0.29 &   -0.47  &   -0.32  &   -0.37  &   -0.24  &   -0.34  &   -0.20  &   -0.26   & \g-0.11  & \g-0.33  &   -0.30  &   -0.46  &   -0.35  &   -0.43  &   -0.39  &   -0.45 \\
 & \textbf{Gyroscope\_covariance}     &  \g0.01  & \g-0.01  & \g0.01   & \g0.01   & \g-0.02  & \g-0.02  & \g0.00   & \g0.03    & \g0.03   & \g-0.01  & \g-0.02  & \g0.00   & \g0.00   & \g0.02   & \g-0.02  & \g0.00  \\
 & \textbf{Point\_filter\_number}     &    0.21  &   0.40   & \g0.06   & \g-0.01  & \g0.05   &   0.17   & \g0.10   &   -0.02   & \g0.10   & \g0.09   &   0.16   &   0.24   & \g0.21   &   0.19   &   0.15   &   0.25  \\
 & \textbf{Filter\_size\_surfaces}    &    0.04  &   0.00   &   0.11   &   0.15   & \g0.14   &   0.24   &   0.25   &   0.51    & \g0.10   & \g0.09   & \g0.16   &   0.29   & \g0.07   &   0.30   &   0.19   &   0.26  \\
 & \textbf{Filter\_size\_map}         &    0.07  & \g0.01   &   0.39   &   0.48   & \g0.29   &   0.32   &   0.32   &   0.40    & \g0.13   &  \g0.65   &   0.28   &   0.37   &   0.29   &   0.51   &   0.41   &   0.40  \\
 & \textbf{LiDAR\_max\_range}         &    -0.42 &   -0.28  & \g-0.10  & \g-0.01  & \g-0.13  &   -0.13  & -0.14    & \g-0.01   & \g-0.10  & \g-0.19  &   -0.19  & \g-0.11  & \g-0.22  &   -0.13  &   -0.15  &   -0.23 \\
\midrule
\multirow{6}{1.2cm}{NTU VIRAL Ouster OS1-16 LiDAR} 
 & \textbf{Acceleration\_covariance} &   0.01   & \g0.01   & \g0.04   & \g0.04   & \g0.09   & \g0.03   & \g-0.01  & \g0.05   & \g0.00   &   0.01   & \g0.00   & \g0.01   & \multicolumn{4}{c}{/} \\
 & \textbf{Gyroscope\_covariance}    & \g0.00   & \g-0.02  & \g0.02   & \g0.00   & \g-0.01  & \g0.00   & \g0.02   & \g0.04   & \g0.00   & \g0.02   & \g0.03   & \g-0.02  & \multicolumn{4}{c}{/} \\
 & \textbf{Point\_filter\_number}    &   0.21   & \g0.18   & \g-0.18  & \g-0.25  & \g-0.27  & \g-0.33  &   0.22   & \g0.07   &   0.56   &   -0.05  & \g-0.09  & \g-0.07  & \multicolumn{4}{c}{/} \\
 & \textbf{Filter\_size\_surfaces}   &   -0.06  & \g-0.07  & \g0.07   & \g-0.31  & \g-0.35  & \g-0.16  & \g-0.03  & \g-0.08  &   0.01   &   -0.19  & \g-0.06  & \g-0.14  & \multicolumn{4}{c}{/} \\
 & \textbf{Filter\_size\_map}        &   -0.07  & \g0.10   & \g0.11   & \g-0.35  & \g-0.01  & \g-0.14  &   -0.06  & \g0.18   &   -0.27  &   0.39   &   0.30   &   0.30   & \multicolumn{4}{c}{/} \\
 & \textbf{LiDAR\_max\_range}        &   0.00   & \g-0.19  & \g0.60   & \g0.26   & \g-0.15  & \g0.16   &   -0.10  & \g-0.09  &   0.18   &   -0.17  &   -0.24  & \g-0.16  & \multicolumn{4}{c}{/} \\
 \bottomrule
\end{tabular}
}
\end{table*}

The \FL2 correlation data (Table \ref{table:fastlio2_correlations}) is used to identify parameters with high need for manual tuning. 
\textbf{LiDAR\_max\_range}, \textbf{Point\_filter\_number}, and \textbf{Filter\_size\_surfaces} each have correlations close to zero in 19/60 influential cases, having the highest need for manual tuning.  
The other parameters (\textbf{Acceleration\_covariance} and \textbf{Filter\_size\_map}) have a small number of cases with weak correlations (5 each), which indicates a low need for manual tuning.

\subsubsection{Correlation data analysis - overarching trends}
\label{subsubsec:fl2_correlation_trends}
\textbf{LiDAR\_max\_range} has negative correlations in 36 out of 38 influential cases in Table \ref{table:fastlio2_correlations}, meaning that setting a low value usually increases ATE, with larger values being preferred. As this prefilter is not included in the original \FL2 implementation (we added it 
to compare its effect with the Cartographer \textbf{LiDAR\_max\_range} filter), we propose to implement it only when satisfactory results are unobtainable by tuning the other parameters.

\textbf{Acceleration\_covariance} has a higher influence on PASTEL sequences (44 cases out of 48) than in NTU VIRAL, where it is influential only on sequence 10. Since the datasets offer the same IMU model (Ouster OS1-16 builtin IMU) we investigated the IMU responses in the two datasets. We found that IMU measurements have less noise in NTU VIRAL than in PASTEL. For instance, on PASTEL sequence 16, the Ouster OS1-16 IMU acceleration has a standard deviation of $1.8\ ms^{-2}$ on the $z$ axis, compared to $0.7\ ms^{-2}$ on the $z$ axis on NTU VIRAL seq. 5. Since the IMU covariance is important for EKF-based LIO processing, this needs to be addressed when tuning LIO algorithms. 
For the Ouster OS1-16 built-in IMU in PASTEL, larger values of \textbf{Acceleration\_covariance} are preferred to cope with the higher standard deviations. 

\textbf{Filter\_size\_map} is the parameter with the most consistent correlations in Table \ref{table:fastlio2_correlations}. As 47/50 influential cases have positive correlations, lower values are preferred. This is expected because this parameter defines the map down-sampling resolution. 
We recommend initially keeping the default value (see Table \ref{table:grid_search}) as reducing the parameter may increase the computational burden in large maps.

\subsubsection{\FL2 tuning recommendations}
\label{subsubsec:fl2_tuning_recommendations}

The above analysis is distilled into the tuning recommendations shown in Table \ref{table:fastlio2_recommendations}, which are guidelines for manual tuning and suggested values for good performance in scenarios similar to the PASTEL and NTU VIRAL datasets, arranged by decreasing need for manual tuning. We recommend to first set the parameters to the recommended values, and to tune them manually if the performance is not satisfactory, starting from the top of the table. Analogously to Cartographer, we expect that a simplified tuning process focusing only on the two most influential parameters to provide satisfactory results, see Section \ref{subsubsec:fl2_validation}.

\begin{table}[pos=ht!]
    \caption{Summary of \FL2 tuning recommendations and recommended starting values. The parameters are arranged by the need for manual tuning in decreasing order.}
    \label{table:fastlio2_recommendations}
    \centering
    \scalebox{0.9}{
    \begin{tabular}{@{\hspace{1mm}}l@{\hspace{1mm}}l}
    \toprule
    \multicolumn{2}{c}{\textbf{\FL2 tuning recommendations}} \\
    \midrule
    \multirow{5}{*}{\textbf{Point\_filter\_number,} }
            & \multirow{5}{5.25cm}{Manually tune starting from default values (\textbf{Point\_filter\_number} = 4 and \textbf{Filter\_size\_surfaces} = 0.5 m). Parameters may interact. Lower values may improve performance. Keep comuptational constraints in mind on low-resource hardware.} \\ \\ \\ 
    \multirow{2}{*}{\textbf{Filter\_size\_surfaces}} \\ \\ \\ \\
            \midrule
    \multirow{5}{*}{\textbf{Acceleration\_covariance}}  
            & \multirow{5}{5.25cm}{Value influenced by IMU properties and mechanical insulation. Use higher value (0.2 m) and consider reducing it if IMU acceleration signal quality is better.} \\ \\ \\ \\ \\
            \midrule
      \multirow{3}{*}{\textbf{Filter\_size\_map}}   
            & \multirow{3}{5.25cm}{Default value (0.5 m) expected to perform well. Low values may increase computational burden on large maps.} \\ \\ \\ 
            \midrule
      \multirow{3}{2.75cm}{\textbf{LiDAR\_max\_range} (added as prefilter)}   
            & \multirow{3}{5.25cm}{
            It may be useful to implement if satisfactory results are not obtainable through other parameters.} 
            \\ \\ \\
            \midrule
      \multirow{3}{*}{\textbf{Gyroscope\_covariance}} 
            & \multirow{3}{5.25cm}{It has low influence in current settings and can be left on its default value (0.1 m).} \\ \\ \\
        \bottomrule
    \end{tabular}
    } 
\end{table}

\subsubsection{Validation of \FL2 tuning recommendations}
\label{subsubsec:fl2_validation}
Analogously to the Cartographer validation , both the recommended values and tuning order from Table \ref{table:fastlio2_recommendations} are validated by applying the recommendations in reverse order (i.e. with rising importances) and observing the ATE distributions for all sequence-LiDAR combinations from \textit{LIO performance data}.

\begin{figure}[pos=ht!]
    \centering
    \includegraphics[width=0.99\linewidth, trim={0cm, 19.0cm, 9.75cm, 0cm}, clip]{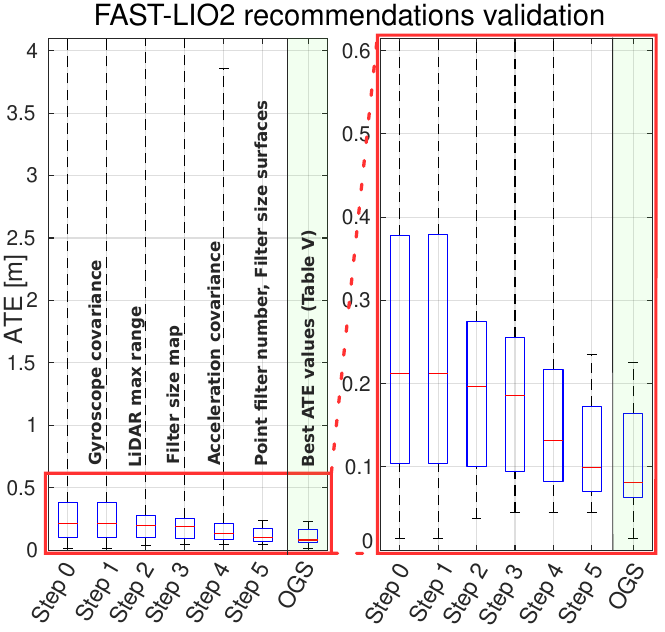}
    \caption{
    ATE distributions considering all sequence-LiDAR combinations in the \emph{LIO performance data} along different validation steps for \FL2. Step 1, which reflects the influence of \textbf{Gyroscope\_covariance}, has minimal influence on the ATE distribution, while the changes in the ATE distributions increase in Steps 2-4, validating the ordering of the parameters in the tuning recommendations. Step 5 manually tunes \textbf{Point\_filter\_number} and \textbf{Filter\_size\_surfaces} for each of the LiDAR and data sequence combinations. As shown, the ATE distribution resulting of the proposed tuning recommendations (Step 5) is similar to the ATE distribution resulting from the optimal grid search, validating that satisfactory performance can be achieved following the proposed recommendations and manually tuning solely two parameters. 
    }
    \label{fig:fastlio2_recommendation_validation}
\end{figure}

The results of the validation process is presented in Figure \ref{fig:fastlio2_recommendation_validation}. Figure \ref{fig:fastlio2_recommendation_validation} validates the ordering of the parameters in Table \ref{table:fastlio2_recommendations}, showing that parameter \textbf{Gyroscope\_covariance}  in Step 1 has a negligible influence on the ATE distributions, with the influence increasing in Steps 2-4. The figure also confirms that applying the proposed simplified tuning process obtains a similar ATE distribution to the distribution of optimal grid search parameters (OGS in the figure), with the extrema, quartiles and medians of the distribution in Step 5 being within 5 cm of their respective points in the OGS distribution.

In addition, Table \ref{table:fastlio2_recommendation_validation} presents the resulting ATE values from the proposed \FL2 tuning recommendations (after Step 5) and compares them to the optimal grid search results for \FL2 from Table \ref{table:best_grid_search_ATEs}. Notably, the tuning procedure obtains results within 5 cm from the grid-search optimal values in 58/60 cases, with the worst ATE being 23.5 cm in NTU VIRAL sequence 8. Hence, we conclude that the \FL2 proposed tuning recommendations are valid for the considered scenarios.

\begin{table}[pos=ht!]
\centering
\caption{\FL2 ATE values for each sequence-LiDAR combination using: i) the parameters resulting from the proposed tuning recommendations (after Step 5), and ii) the grid search optimal parameters (shown in Table \ref{table:best_grid_search_ATEs}, repeated here for ease of comparison). All values are in centimeters.
}
\label{table:fastlio2_recommendation_validation}
\scalebox{0.9}{
\begin{tabular}{l|c@{\hspace{2mm}}c|c@{\hspace{2mm}}c|c@{\hspace{2mm}}c}
\toprule
\multicolumn{7}{c}{\textbf{Individual ATE values from \FL2 validation Step 5 }} \\
\multicolumn{7}{c}{\textbf{compared to optimal values from Table \ref{table:best_grid_search_ATEs} [cm]}} \\ 
\midrule
\mrIII{\textbf{Data sequence}} & \multicolumn{2}{c|}{\textbf{Velodyne}}  & \multicolumn{2}{c|}{\textbf{Ouster}} & \multicolumn{2}{c}{\textbf{Ouster}} \\
                      & \multicolumn{2}{c|}{\textbf{VLP16}}     & \multicolumn{2}{c|}{\textbf{OS1-16}} & \multicolumn{2}{c}{\textbf{OS0-128}} \\
             & \textbf{Step 5} & \textbf{OGS} & \bf{Step 5} & \textbf{OGS} & \bf{Step 5} & \bf{OGS} \\   
\midrule              
PASTEL    1  &     6.5     &  6.4     &  10.0  &  1.4    &  6.5     &  6.3    \\
PASTEL    2  &     7.5     &  7.5     &  10.4  &  10.3   &  8.8     &  6.9    \\
PASTEL    3  &     7.0     &  5.6     &  8.4   &  7.9    &  6.7     &  6.1    \\
PASTEL    4  &     10.9    &  7.1     &  14.2  &  9.4    &  10.3    &  7.4    \\
PASTEL    5  &     6.1     &  4.8     &  7.3   &  7.2    &  6.4     &  6.3    \\
PASTEL    6  &     5.5     &  4.6     &  7.3   &  6.9    &  4.5     &  4.2    \\
PASTEL    7  &     5.9     &  5.5     &  10.0  &  9.8    &  8.4     &  7.7    \\
PASTEL    8  &     5.4     &  5.3     &  8.1   &  8.0    &  5.1     &  3.7    \\
PASTEL    9  &     6.1     &  6.1     &  8.2   &  8.1    &  6.2     &  5.4    \\
PASTEL    10 &     7.0     &  6.1     &  8.2   &  8.1    &  5.8     &  5.5    \\
PASTEL    11 &     8.4     &  8.2     &  16.4  &  16.3   &  14.2    &  14.0   \\
PASTEL    12 &     10.9    &  10.4    &  18.9  &  18.3   &  14.8    &  12.8   \\
PASTEL    13 &     9.2     &  7.9     &  16.9  &  16.5   &  15.6    &  15.0   \\
PASTEL    14 &     10.0    &  9.9     &  16.6  &  16.6   &  17.6    &  13.4   \\
PASTEL    15 &     9.5     &  9.5     &  18.5  &  18.5   &  15.2    &  13.6   \\
PASTEL    16 &     6.4     &  6.4     &  13.2  &  13.0   &  9.3     &  7.3    \\
\midrule                                           
NTU VIRAL 1  & \multicolumn{2}{c|}{/} &  21.2  & 20.3    & \multicolumn{2}{c}{/} \\
NTU VIRAL 2  & \multicolumn{2}{c|}{/} &  20.4  & 18.7    & \multicolumn{2}{c}{/} \\
NTU VIRAL 3  & \multicolumn{2}{c|}{/} &  22.7  & 20.7    & \multicolumn{2}{c}{/} \\
NTU VIRAL 4  & \multicolumn{2}{c|}{/} &  21.7  & 20.5    & \multicolumn{2}{c}{/} \\
NTU VIRAL 5  & \multicolumn{2}{c|}{/} &  21.6  & 20.4    & \multicolumn{2}{c}{/} \\
NTU VIRAL 6  & \multicolumn{2}{c|}{/} &  21.7  & 19.5    & \multicolumn{2}{c}{/} \\
NTU VIRAL 7  & \multicolumn{2}{c|}{/} &  23.2  & 22.6    & \multicolumn{2}{c}{/} \\
NTU VIRAL 8  & \multicolumn{2}{c|}{/} &  23.5  & 22.3    & \multicolumn{2}{c}{/} \\
NTU VIRAL 9  & \multicolumn{2}{c|}{/} &  23.1  & 21.5    & \multicolumn{2}{c}{/} \\
NTU VIRAL 10 & \multicolumn{2}{c|}{/} &  19.5  & 19.0    & \multicolumn{2}{c}{/} \\
NTU VIRAL 11 & \multicolumn{2}{c|}{/} &  19.1  &  7.5    & \multicolumn{2}{c}{/} \\
NTU VIRAL 12 & \multicolumn{2}{c|}{/} &  19.9  & 19.6    & \multicolumn{2}{c}{/} \\
\bottomrule    
\end{tabular}    
} 
\end{table}    

\section{Conclusion and future work}
\label{sec:conclusion}

The performance of LiDAR-based LIO and SLAM algorithms is strongly dependent on the scenario, robot motion and LiDAR characteristics, often requiring intensive tuning processes. This paper analyzes the influence of Cartographer and \FL2 parameters on the estimated trajectory accuracy in low-to-medium altitude quadrotor flights. We aim to decrease the tuning effort by increasing the understanding of the LIO algorithm parameters, narrowing the parameter search space, and facilitating the tuning process.

The analysis starts with identifying parameters of interest and defining values for each parameter which can be considered low, medium, and high settings. All parameter combinations were evaluated on two public datasets which cover three different LiDAR models in diverse environments and with diverse flight characteristics. The parameter influence was analyzed using Pearson's correlation to determine the relationship between the parameter values and resulting ATE metrics, while the magnitude of parameters' influence was determined using random forest permutation importances from random forest regressors trained on the obtained LIO algorithm performance data.

As a result of these analyses, this paper provides and validates tuning recommendations for the analyzed parameters, which provide practitioners with recommended starting tuning values and a proposed tuning order. The paper proposes a simplified tuning process focusing on the two most influential parameters of each algorithm. Using the proposed tuning recommendations, both LIO algorithms obtain on the analyzed datasets ATE values within 5 cm to the optimal performance found in the grid search procedure in 94\% of the analyzed cases.

This work opens a wide field for future research. One of them is to expand it by analyzing possibilities of parameter interactions, which were not considered in this work. The extension to analyze SLAM (instead of LIO) parameters is object of current research. This work focused on lower-altitude flights. The extension to high altitude flight scenarios for applications such as surveillance and mapping is object of current research. Finally, the development of methods to relate the aerial LiDAR data characteristics (e.g., flight altitude, velocity, distances from obstacles) with the choice of algorithmic parameters would enable a system that could recommend parameter sets for a given application just by analyzing the data collected in a test flight.

\printcredits

\section*{Acknowledgements}

The work presented in this paper was funded by the European Union under the MARBLE project (GA No: 101136349) and through the National Recovery and Resiliance Plan under the grant NPOO.C3.2.R3-I1.01.0274 (project OTIP - Point cloud for industrial plant digitalization) and by MICIU/AEI/10.13039/501100011033 and ERDF, EU, through the Project "RAISE: Robots A\'ereos Inteligentes en Cooperaci\'on Estrecha con Sistemas IoT para la Inspecci\'on Avanzada de Viaductos", under Grant PID2023-149683OB-I00. Views and opinions expressed are however those of the author(s) only and do not necessarily reflect those of the European Union or the European Research Executive Agency. Neither the European Union nor the granting authority can be held responsible for them.

\bibliographystyle{model1-num-names}

\bibliography{bibliography}

\bio{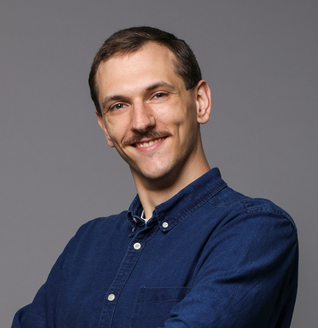}
\textbf{Robert Milijas} 
received the B.S. and M.S. degrees in electrical engineering and information technology from the University of Zagreb, Croatia, in 2017 and 2019. From 2019 until 2025 he was a Research Assistant with the University of Zagreb Faculty of Electrical Engineering and Computing Laboratory for Robotics and Intelligent Control Systems (LARICS) where he is currently pursuing the PhD degree in electrical engineering. Since 2025 he is a Junior Researcher at the CoE MARBLE - Centre of Excellence in Maritime Robotics and Technologies for Sustainable Blue Economy. He is the author of two journal articles and more than eight conference papers. His research interests include SLAM, unmanned aerial vehicles and LiDAR perception. 
\endbio

\bio{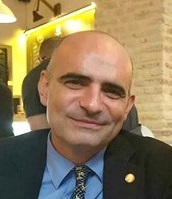}
\textbf{J. Ramiro Mart\'inez-de Dios} Full professor in robotics and Director of the Innovation Center for Unmanned Aerial Vehicles and Unmanned Air Mobility of the University of Seville. His research interests include robot perception, multi-robot cooperation, robot localization and mapping, and sensor fusion, where he has authored  $>$170 publications, has coordinated $>$25 R\&D projects and $>$20 technology transfer contracts. He has received 5 international awards, is member of the Editorial Board of $>$5 journals, Associated Editor of IEEE RA-L, and member of the Board of Directors of euRobotics.
\endbio

\balance

\bio{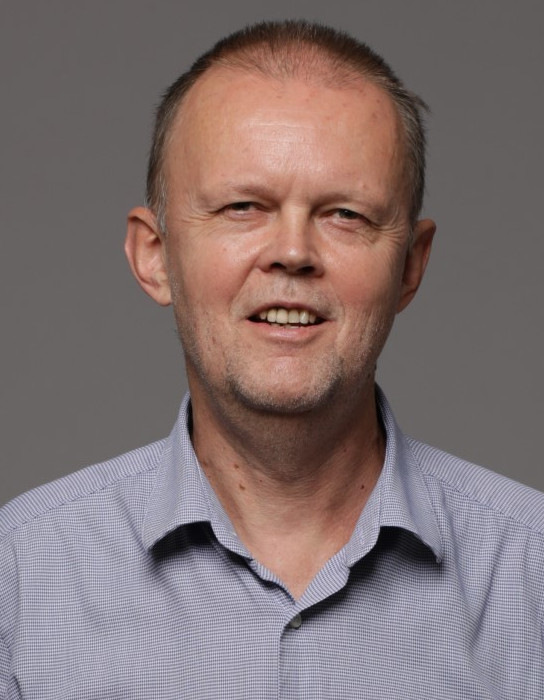}
\textbf{Stjepan Bogdan }is currently a Full Professor with the Laboratory for Robotics and Intelligent Control Systems (LARICS), Faculty of Electrical Engineering and Computing, University of Zagreb. He is the coauthor of four books and has published more than 200 conferences and journal articles. He was the Principal Investigator and a Researcher on 30 national and international scientific projects. His research interests include autonomous systems, aerial robotics, multi-agent systems, intelligent control systems, bio-inspired systems, and discrete event systems. He serves as an associate editor for several scientific journals and conferences and was a program and organizing committee member of major control and robotics conferences.
\endbio

\end{document}